\documentclass[sigconf]{acmart}

\AtBeginDocument{%
  \providecommand\BibTeX{{%
    \normalfont B\kern-0.5em{\scshape i\kern-0.25em b}\kern-0.8em\TeX}}}

\copyrightyear{2020} 
\acmYear{2020} 
\setcopyright{acmcopyright}
\acmConference[KDD '20]{Proceedings of the 26th ACM SIGKDD Conference on Knowledge Discovery and Data Mining}{August 23--27, 2020}{Virtual Event, CA, USA}
\acmBooktitle{Proceedings of the 26th ACM SIGKDD Conference on Knowledge Discovery and Data Mining (KDD '20), August 23--27, 2020, Virtual Event, CA, USA}
\acmPrice{15.00}
\acmDOI{10.1145/3394486.3403379}
\acmISBN{978-1-4503-7998-4/20/08}
\usepackage{siunitx}
\usepackage{subcaption}
\usepackage{caption}
\usepackage{bbold}
\usepackage{amsmath,amssymb}
\usepackage{amsmath}
\usepackage{enumitem}
\usepackage{multicol}
\usepackage{verbatim}
\usepackage{bbm}
\usepackage{dsfont}
\usepackage{bbold}
\usepackage{csquotes}
\MakeOuterQuote{"}
\begin{document}
\setlist[enumerate]{leftmargin=*}

\title{An Empirical Analysis of Backward Compatibility \\in Machine Learning Systems}

\author{Megha Srivastava}
\affiliation{%
   \institution{Microsoft Research}}

\author{Besmira Nushi}
\affiliation{%
  \institution{Microsoft Research}}

\author{Ece Kamar}
\affiliation{%
  \institution{Microsoft Research}}

\author{Shital Shah}
\affiliation{%
  \institution{Microsoft Research}}

\author{Eric Horvitz}
\affiliation{%
  \institution{Microsoft Research}}

\renewcommand{\shortauthors}{Srivastava et al.}

\begin{abstract}
In many applications of machine learning (ML), updates are performed with the goal of enhancing model performance. However, current practices for updating models rely solely on isolated, aggregate performance analyses, overlooking important dependencies, expectations, and needs in real-world deployments. We consider how updates, intended to improve ML models, can introduce new errors that can significantly affect downstream systems and users. For example, updates in models used in cloud-based classification services, such as image recognition, can cause unexpected erroneous behavior in systems that make calls to the services. Prior work has shown the importance of "backward compatibility" for maintaining human trust. We study challenges with backward compatibility across different ML architectures and datasets, focusing on common settings including data shifts with structured noise and ML employed in inferential pipelines. Our results show that (i) compatibility issues arise even without data shift due to optimization stochasticity, (ii) training on large-scale noisy datasets often results in significant decreases in backward compatibility even when model accuracy increases, and (iii) distributions of incompatible points align with noise bias, motivating the need for compatibility aware de-noising and robustness methods. 
\end{abstract}

\maketitle
\section{Introduction}
\label{section:intro}
Enthusiasm around applying machine learning (ML) methods in high-stakes domains such as healthcare and transportation is balanced by concerns about their reliability. Prior works on ML reliability and robustness have sought to develop techniques that enable models to  perform successfully under the presence of data shifts or adversarial perturbations~\cite{Cheplygina2018Copd, Wen2014Test,Kurakin2016AdversarialML,Madry2017TowardsDL}. 
However, these studies investigate the reliability of models in isolation, quantified only by aggregate performance metrics. In this work, we focus on challenges that arise when models are employed within larger systems, including (i) pipelines composed of multiple components, and in (ii) human-machine interaction. In both cases, dependencies and expectations about model behavior---and the influences of changes on the performance of the larger systems they compose---must be considered during model updates. For example, \emph{when we ``improve'' the overall performance of an existing model, are there any hidden costs to the gain in accuracy? What new errors and failures are introduced that did not exist previously, thus decreasing reliability?}

We show how practitioners can leverage the notion of \emph{backward compatibility} to answer these questions. Backward compatibility in ML systems was first introduced by Bansal et. al~\cite{bansal2019updates} to describe the phenomenon of partial model regress in the context of  human-AI collaboration. They observe that updated models that are more accurate may still break human expectations and trust when introducing errors that were not present in earlier versions of the model. However, considerations about backward compatibility are not limited to human-AI collaboration settings; new, unexpected errors induced by updates aimed at refining models pose problems with reliability when the models are used in larger, integrative AI systems composed of multiple models or other computing components that work in coordination. 

Furthermore, the phenomenon of how and when costly backward incompatibility occurs in a learning context is not very well understood. To the best of our knowledge, no work to date has empirically investigated the causes of compatibility failures in machine learning, and the extent of this issue across datasets and models. We seek to investigate backward compatibility through empirical studies and demonstrate how designers of ML systems can take backward compatibility into consideration to build and maintain more reliable systems.

\begin{figure}
    \includegraphics[width=0.45\textwidth]{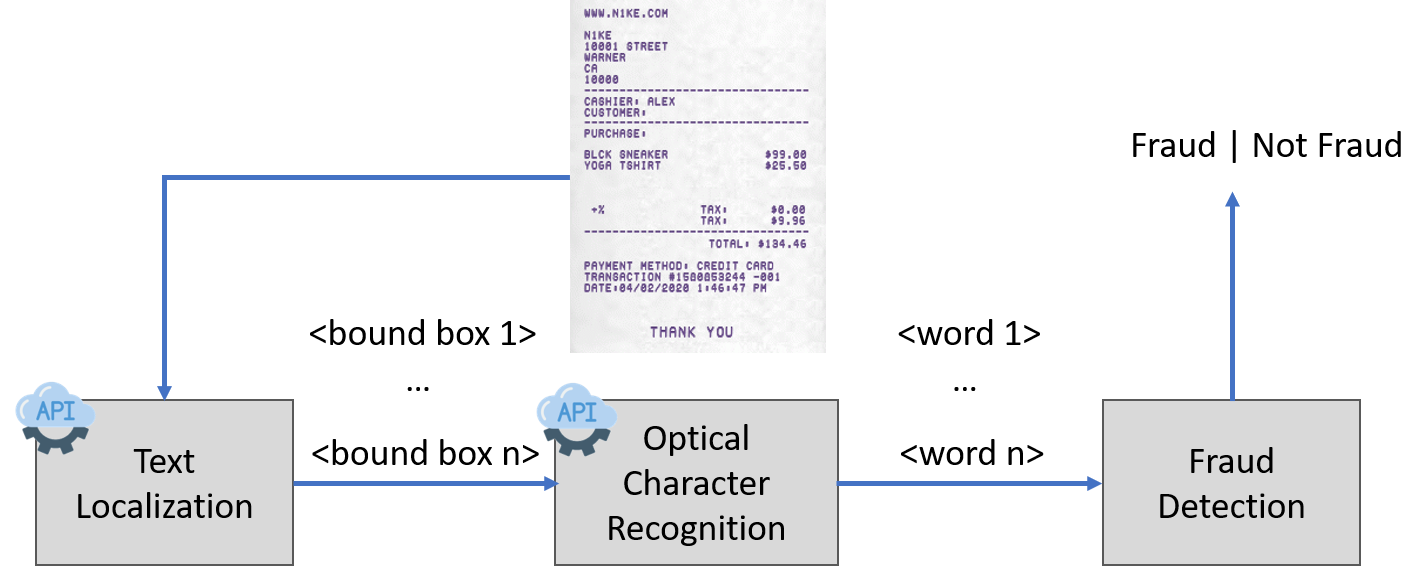}
    \caption{Example of a machine learning pipeline.}
    \Description{Example of a machine learning pipeline showing importance of backward compatibility}
    \label{fig:intro}
\vspace{-4mm}
\end{figure}
  
As a motivating example (Figure~\ref{fig:intro}), consider the financial department of a company that processes expense reports to detect receipt fraud. The department may rely on an off-the-shelf optical character recognition (OCR) model to process receipts \cite{kruppa2008patent}.  Over time, the team may develop and encode a blacklist of common spoofed company names found to be fraudulent, such as ``G00gle llc.'' or ``Nlke''. In this case, the financial services team builds the heuristic blacklist component with expectations on the OCR's ability to detect these names. Moreover, the heuristic rules have been optimized to perform well in accordance with the original OCR model used in development. 

Meanwhile, in pursuing improvement in the OCR's overall performance, engineers may update the training data with a larger and perhaps noisier dataset from a variety of handwritten text sources. Engineers may celebrate increases in model accuracy and generalizability. However, if the update decreases the OCR performance on specific characters present in the blacklisted words, the financial department may experience costly failures when fraudulent receipts now go undetected, despite believing that they are using a better model. For example, if the newly added dataset contains biased noise and frequently mislabels ``0'' digits as ``o'' characters, the word ``G00gle llc.'' may be wrongly recognized as ``Google llc.'', which is not in the blacklist. Similar scenarios in high-stakes settings such as healthcare can lead to even more serious consequences. 

Motivated by potential downstream errors such as the one above, we emphasize that machine learning practitioners should consider potential backward compatibility issues before they decide to update models. Such practice is similar to common practices that software engineers follow when modifying traditional software~\cite{bosch2009software}. More specifically, we make the following contributions:
\begin{enumerate}
    \item We expand the empirical understanding of when and how backward compatibility issues arise in machine learning and the relationship to example forgetting during retraining.
    \item We characterize backward compatibility under different noise settings using a variety of datasets (tabular, vision, language) and model architectures (linear, CNN, ResNet, BERT).
    \item We illustrate and discuss backward compatibility from the perspective of maintaining and monitoring the performance of modularized ML pipelines.
    \item We highlight how ML practitioners can use the presented results and methodology to create best practices and tools for updating and diagnosing learning models.
\end{enumerate}

 The rest of the paper is organized as follows: Section~\ref{section:relatedwork} positions the paper in the context of related work. Section~\ref{section:problem} defines the setting of model updates and the backward compatibility metrics we use to characterize the phenomenon. Section~\ref{section:method} details the experimental setup. Section~\ref{section:stoch} studies the effect of optimization stochasticity on backward compatibility as a baseline, before studying backward compatibility in the presence of noisy model updates in Section~\ref{section:exp}. Finally, Section~\ref{section:pipeline} shows how backward compatibility analyses can help to identify failures in ML pipelines.

\section{Background and Related Work}
\label{section:relatedwork}

\noindent \textbf{Backward compatibility in Machine Learning.} Bansal et. al~\cite{bansal2019updates} introduced backward compatibility in the context of preserving human expectations and trust, and proposed a loss function to penalize newly introduced errors, which we further study in Section 6.3. Our work  expands the understanding of backward compatibility in ML, and we add a new perspective by highlighting backward compatibility challenges that arise when inferences from a retrained model are used by other components of a larger system. 

The challenge of understanding how changes in one component propagate through an entire system has attracted recent interest. ~\cite{sculley2015hidden,nushi2017human,andrist2017went} discuss how hidden data and feature dependencies can cause component entanglement, unexpected system behavior, and downstream performance regression~\cite{nushi2017human,andrist2017went}. Such issues have also been highlighted in surveys on software engineering challenges in ML~\cite{amershi2019software,zhang2019empirical}. New  performance evaluation~\cite{chung2019slice,nushi2018towards,zhang2018manifold} approaches suggest reporting performance on data slices rather than aggregate measures that overlook partial model regression. Our work advises that measures of backward compatibility should be included when monitoring model performance during updates.

\noindent \textbf{Catastrophic forgetting and transfer learning.} Backward compatibility is relevant to the phenomenon of catastrophic forgetting~\cite{french1999catastrophic,mccloskey1989catastrophic}, which refers to situations where a learning model forgets old tasks while being trained in a sequential fashion to solve new tasks. Literature on catastrophic forgetting~\cite{kemker2018measuring,goodfellow2013empirical} highlights ``learning interference'' as a cause for forgetting in models with a fixed capacity. In contrast, we investigate fluctuations that do not involve changes in the task or concept definition. Despite this simplification, we find retrained models are still subject to forgetting individual or similar examples.~\cite{toneva2018empirical} present settings where the task definition has not changed, and focus on understanding cases of \emph{example forgetting}. In these events, individual examples may be forgotten (multiple times) by the model during gradient-based optimization. The authors find that the most ``forgotten'' examples are often the most difficult, such as atypical inputs and noisy labels. We connect these results with our findings in Section \ref{section:stoch}, where we show how example forgetting events relate to model incompatibility. 

\noindent \textbf{Data cleaning and distributional shifts.} Data cleaning is a topic of long-term interest in the data management and mining community~\cite{
rahm2000data,DBLP:journals/pvldb/RekatsinasCIR17,DBLP:conf/sigmod/DallachiesaEEEIOT13}. Techniques addressing problems like outlier detection and data denoising are often agnostic to machine learning models, as the cleaning process is not guided by the influence on a particular model. Although model-agnostic data cleaning is important for cleaning generic data repositories consumed by multiple and sometimes unknown models, it is important to map changes in data characteristics to model performance to fully understand the impact of data quality when updating an already deployed model. While we do not propose a new technique for data cleaning, our findings can pave the way for future techniques that preserve the performance of systems in the presence of data shifts and noise.  

Other studies have outlined the influence of class label noise~\cite{frenay2013classification,krause2016unreasonable} and feature noise~\cite{hendrycks2019benchmarking} on model behavior, informing a parallel line of work on increasing model robustness to noise~\cite{Sukhbaatar2014TrainingCN,Natarajan2013LearningWN,Li2017LearningFN,Jiang2017MentorNetLD}. While these works show that modern neural networks can be somewhat robust to both label and feature noise (in non-adversarial settings), their evaluation heavily relies on aggregate performance metrics and does not investigate the impact of biased noise that affects only certain data clusters, as often occurs in the real world.

\section{Problem Set-Up}
\label{section:problem}
In both traditional software and machine learning systems, backward compatibility is a reliability concern emerging from \textit{model updates}. There are many possible incentives for updating a model, including re-training on a larger, more diverse, and perhaps noisier dataset to increase the model's generalizability and  overall accuracy. In this section, we describe this specific setting of "noisy model updates," and then define two measures of backward compatibility.
\vspace{-4mm}
\subsection{Noisy Model Updates}
Consider a machine learning team that has deployed the first version of a model, which was carefully trained on a small and clean dataset. While the small data quantity enabled the team to verify all labels and ensure good data quality, the team may now wish to improve model performance on a greater variety of inputs by using a larger and more diverse dataset. Due to high financial cost and time, the team may resort to noisier approaches to create this large dataset, such as crowdsourcing labels~\cite{law2011human} or using weak supervision~\cite{ratner2019snorkel}, a popular alternative due to its automation opportunities. Some examples include scraping web data \cite{Mendels2015Speech}, leveraging social media tags as labels, or collecting data via image search. 

We refer to this practice as a \textit{Noisy Model Update}, which has been performed across a variety of tasks, including language and vision domains, and has been shown to improve overall accuracy~\cite{cheplygina2019not,ratner2019snorkel}. However, the additional data may also contain biased noise, concentrated in particular regions or classes of the dataset. Imagine a team is collecting data from social media to enrich the dataset. In social media, some keywords may be ambiguous and contain examples from two or more classes at the same time. For example, the keyword "book" on Instagram shows results for books, inspirational quotes, and travel destinations. Model updates with data including these biased concentrations of noise may harm backward compatibility, necessitating analysis of the ways the newer model may be unreliable despite its increased accuracy. In Section \ref{section:exp}, we analyze backward compatibility with respect to three different types of noise: label noise, feature noise, and outlier noise.

We are mainly interested in cases when  noise does not affect the whole dataset uniformly because uniform noise distribution is more likely to impact the overall accuracy, reducing the incentive for a model update. Nevertheless, as we show in Section~\ref{section:pipeline}, even for uniform noise, some classes are more affected than others as they may be less distinguishable.

\noindent \textbf{Problem Definition.} More formally, let $h_1 =$ Model 1, $D_1 =$ Training Dataset 1, $h_2 =$ Model 2, and $D_2 =$ Training Dataset 2, where $h_1$ is trained on $D_1$, and $h_2$ is trained on $D_2$. Each model h is a function $h: x \rightarrow y$, where $x$ is an input instance and $y$ is the ground truth label for the input. Backward compatibility of $h_2$ with regards to $h_1$ is evaluated on a separate held-out test set $D_{\text{test}}$. One can view $D_{\text{test}}$ as a clean dataset that the team uses for performance evaluation in general. It is typically chosen such that its distribution resembles the expected real-world distribution, but  in practice the resemblance varies by a large margin. Further, our problem definition makes the following assumptions to control for other factors that may affect backward compatibility: (1) $h_1$ and $h_2$ have the same model architecture; (2) the set of possible labels in $D_1$ and $D_2$ are the same; (3) the classification accuracy of $h_2$ on $D_{\text{test}}$ is higher than the accuracy of $h_1$ on $D_{\text{test}}$, motivating the model update.

In our experiments, we first simplify the problem to set an upper-bound baseline on backward compatibility by keeping $D_1=D_2$ (Section~\ref{section:stoch}). In this case, any discrepancies between $h_1$ and $h_2$ are due to optimization stochasticity. Afterwards, we proceed with results for the noisy update problem in Section \ref{section:exp}, where $D_1 \subset D_2$. 

\subsection{Measuring Backward Compatibility}
  \begin{figure}
    \includegraphics[width=0.75\columnwidth]{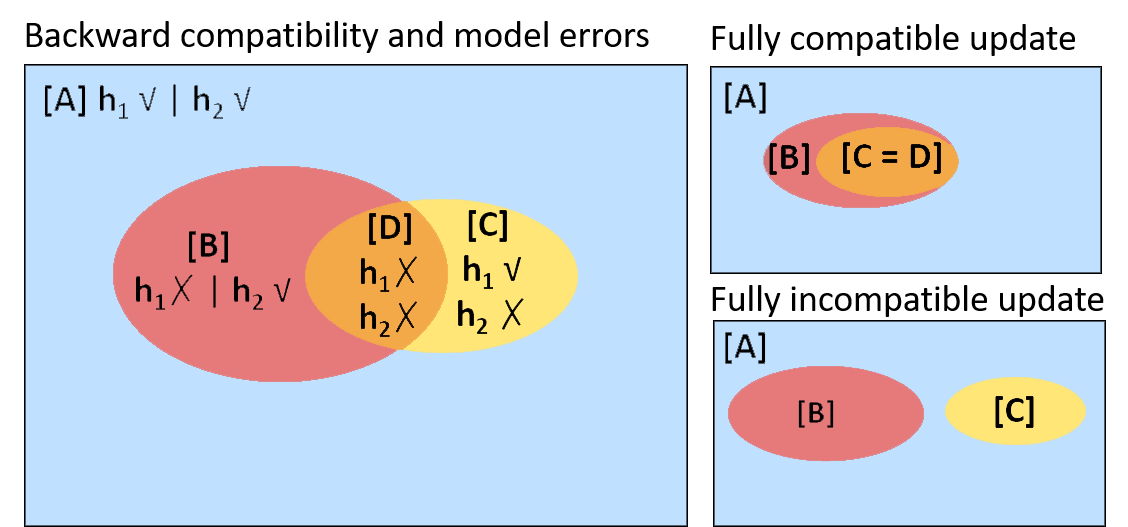}
    \caption{Backward compatibility as described by model errors. In fully compatible updates, the updated model ($h_2$) only refines errors of the previous model ($h_1$). In fully incompatible updates, the errors of the two models are disjoint. $\text{BTC}= 1 - \frac{\text{A} + \text{B}}{\text{A} + \text{C}}$, $\text{BEC}= 1 - \frac{\text{C}}{\text{D} + \text{C}}$.}
    \Description{Insert Description}
    \label{fig:metrics}
\vspace{-4mm}
\end{figure}
We define two measures of backward compatibility, which we use in our experiments (See Figure \ref{fig:metrics}). The first is the \textbf{Backward Trust Compatibility (BTC) score}, which is the ratio of points in a held-out test set (e.g., $D_{\text{test}}$) that $h_2$ predicted correctly among all points $h_1$ had already predicted correctly.

\begin{equation}
\begin{split}
  \text{BTC} = \frac{\sum_{i=1}^{|D|} \mathds{1}[h_1(x_i) = y_i, h_2(x_i) = y_i]}{\sum_{i=1}^{|D|} \mathds{1}[h_1(x_i) = y_i]}
  \end{split} 
\label{eq:btc}
\end{equation}

BTC matches the backward compatibility score in \citep{bansal2019updates} and semantically describes the percentage of trust that is preserved after the update. 
However, a challenge with  measuring only BTC is that if the updated model $h_2$ achieves almost-perfect accuracy, such as when applying deep neural networks to extremely low sample-complexity problems (Section \ref{section:exp}), BTC may be high even if most errors caused by $h_2$ are new and unexpected. We thus consider a second metric, the \textbf{Backward Error Compatibility (BEC)} score, which is the proportion of points in a held-out test set that $h_2$ predicted incorrectly, out of which $h_1$ also predicted incorrectly, thus capturing the probability that a mistake made by $h_2$ is not new. 
\begin{equation}
\begin{split}
  \text{BEC} = \frac{\sum_{i=1}^{|D|} \mathds{1}[h_1(x_i) \neq y_i, h_2(x_i) \neq y_i]}{\sum_{i=1}^{|D|} \mathds{1}[h_2(x_i) \neq y_i]}
  \end{split} 
\label{eq:bec}
\end{equation}
 These unexpected errors may negatively impact downstream components in a pipeline that may have learned to suppress the initial errors of $h_1$ or are mitigating them via traditional error handling and heuristics. We propose that developers of ML models should measure both \textbf{BTC} and \textbf{BEC} when considering a model update, rather than solely improvement in accuracy.

\section{Experimental Methods}
\label{section:method}
Our empirical evaluations cover several tasks in increasing order of dataset and model complexity. As an upper-bound baseline on backward compatibility, we first focus on settings without a model update (where data sets are identical through retraining, $D_1 = D_2$). To show that backward compatibility issues exist even in simple models, we apply a logistic regression model on a FICO binary-classification task (Section \ref{section:stoch}). We then investigate the effect of different types of noise on (1) the MNIST digit classification task with a 3-layer CNN, (2) the more complex CIFAR-10 image recognition task using a ResNet-18 model, and then (3) fine tuning the state-of-the-art  BERT language model for an IMDB sentiment analysis task (Section \ref{section:exp}). Finally, we analyze downstream errors in a ML pipeline where one component applies a 3-layer CNN on the Chars74K Character Recognition task (Section \ref{section:pipeline}). 

\noindent \textbf{Experiment Design.} For all experiments, we train two models $h_1$ and $h_2$ on datasets $D_1$ and $D_2$ respectively, where $D_1 = D_2$ for the optimization stochasticity baselines and $D_1 \subset D_2$ for experiments involving model updates. We evaluate the accuracy of $h_1$ and $h_2$ on the same held-out test set, on which we also calculate the BTC and BEC scores as $h_2$'s backward compatibility with respect to $h_1$.  For experiments investigating the effect of noise, such as those in Figure \ref{fig:test}, we calculate the \textit{gain in test accuracy} from $h_2$, and the BTC and BEC scores over varying noise amounts, noting that in practice a developer likely uses a dataset with an unknown fixed amount of noise. Finally, we analyze the set of \textit{incompatible points} - points that $h_1$ predicted correctly yet $h_2$ missed - to gain insight on the types of points the new model $h_2$ is unreliable on. 

\noindent \textbf{Model and Dataset Details.} For all datasets, we aimed to use a close to state-of-the-art models that has the most accessible implementation, as our target audience is ML practitioners. Below, we list all model and dataset details, in order of their appearance:

\begin{enumerate}
    \item \textbf{FICO Credit Score Risk Classification} \cite{fico2020data}: \textit{\# Classes}: 2, \textit{Dataset Sizes}: $|D_1|=|D_2|= 6000$, $|D_{test}|= 1973$, \textit{Model}: Logisitic Regression, \textit{Learning Rate}: $1e-4$, \textit{ Train Epochs}: 100 
    \item \textbf{MNIST Digit Classification} \cite{lecun2010mnist}: \textit{\# Classes}: 10, \textit{Dataset Sizes}: $|D_1|=4800, |D_2|= 48000$, $|D_{test}|= 12,000$, \textit{Model}: 3-Layer CNN, \textit{Learning Rate}: $1e-2$, \textit{Train Epochs}: 50 
    \item \textbf{CIFAR-10 Object Recognition} \cite{Krizhevsky09learningmultiple}: \textit{\# Classes}: 10, \textit{Dataset Sizes}: $|D_1|=10000, |D_2|= 50000$, $|D_{test}|= 10000$, \textit{Model}: ResNet-18, \textit{Learning Rate}: $0.1$, \textit{ Train Epochs}: 35 
    \item \textbf{IMDB Movie Review Sentiment Analysis} \cite{maas2011imdb}: \textit{\# Classes}: 2, \textit{Dataset Sizes}: $|D_1|=400, |D_2|= 1600$, $|D_{test}|= 400$, \textit{Model}: Finetuned BERT-Base Language Model, \textit{Learning Rate}: $4e-5$, \textit{ Train Epochs}: 2
    \item \textbf{Chars74K Char. Recognition (OCR Pipeline)} \cite{deCampos09}: \textit{\# Classes}: 62, \textit{Dataset Size}: $|D_1|=4960, |D_2|= 24800$, $|D_{test}|= 6200$, \textit{Model}: 3-Layer CNN, \textit{Learning Rate}: $1e-4$, \textit{Train Epochs}: 20
\end{enumerate}We train all models using stochastic gradient descent (SGD) and ensure all classes are equally represented in the training dataset. 

\section{Backward Compatibility and Optimization Stochasticity}
\label{section:stoch}
\begin{figure}[t]
    \begin{subfigure}[t]{0.5\linewidth}
    \centering
    \includegraphics[width=\linewidth]{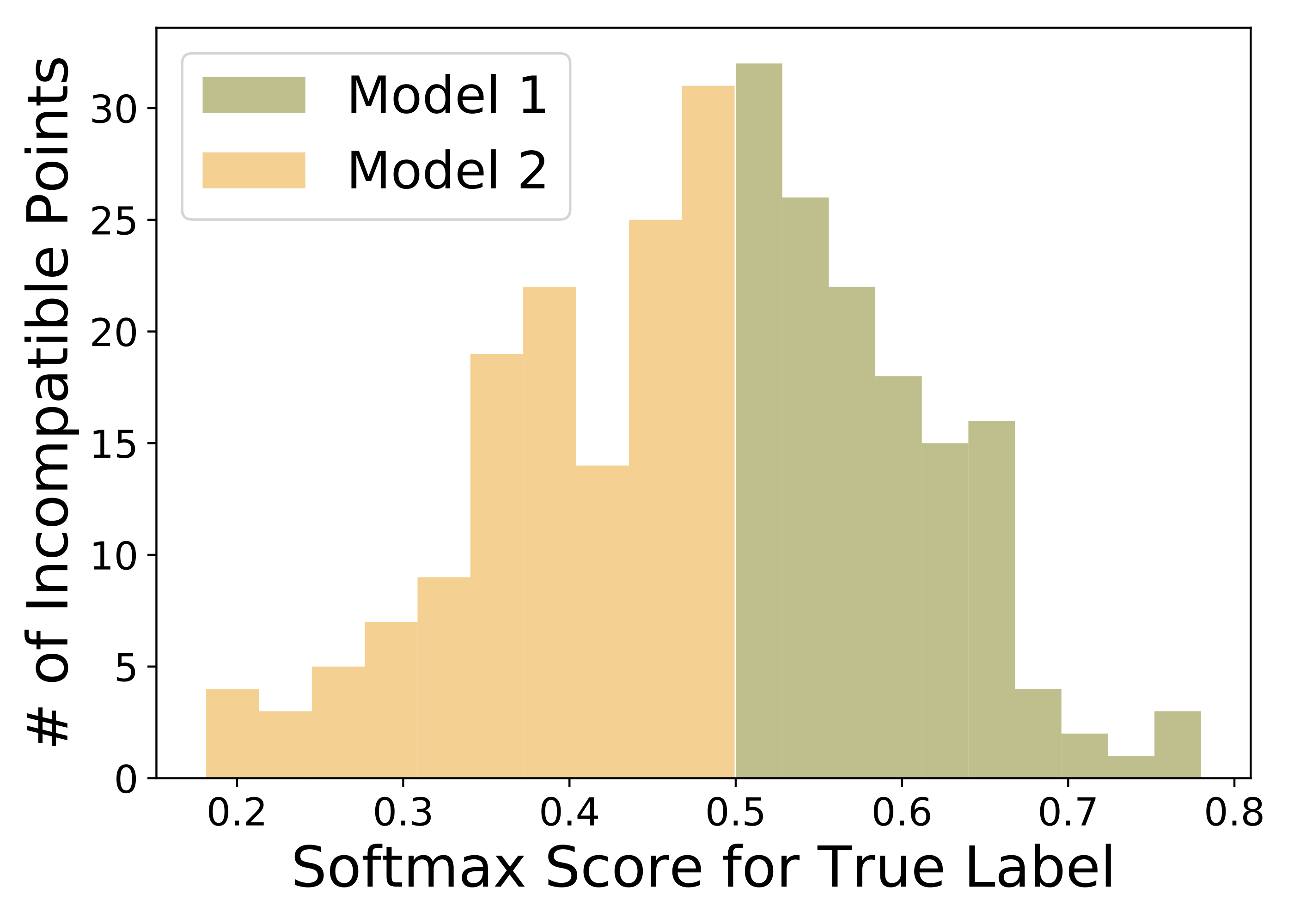}
    \caption{}
    \label{fig:confidence}
    \Description{Distribution of incompatible examples over the output softmax score for two individual models.}
    \end{subfigure}%
    ~\hspace{-15pt}
    \hfill
    \begin{subfigure}[t]{0.5\linewidth}
    \centering
    \includegraphics[width=\linewidth]{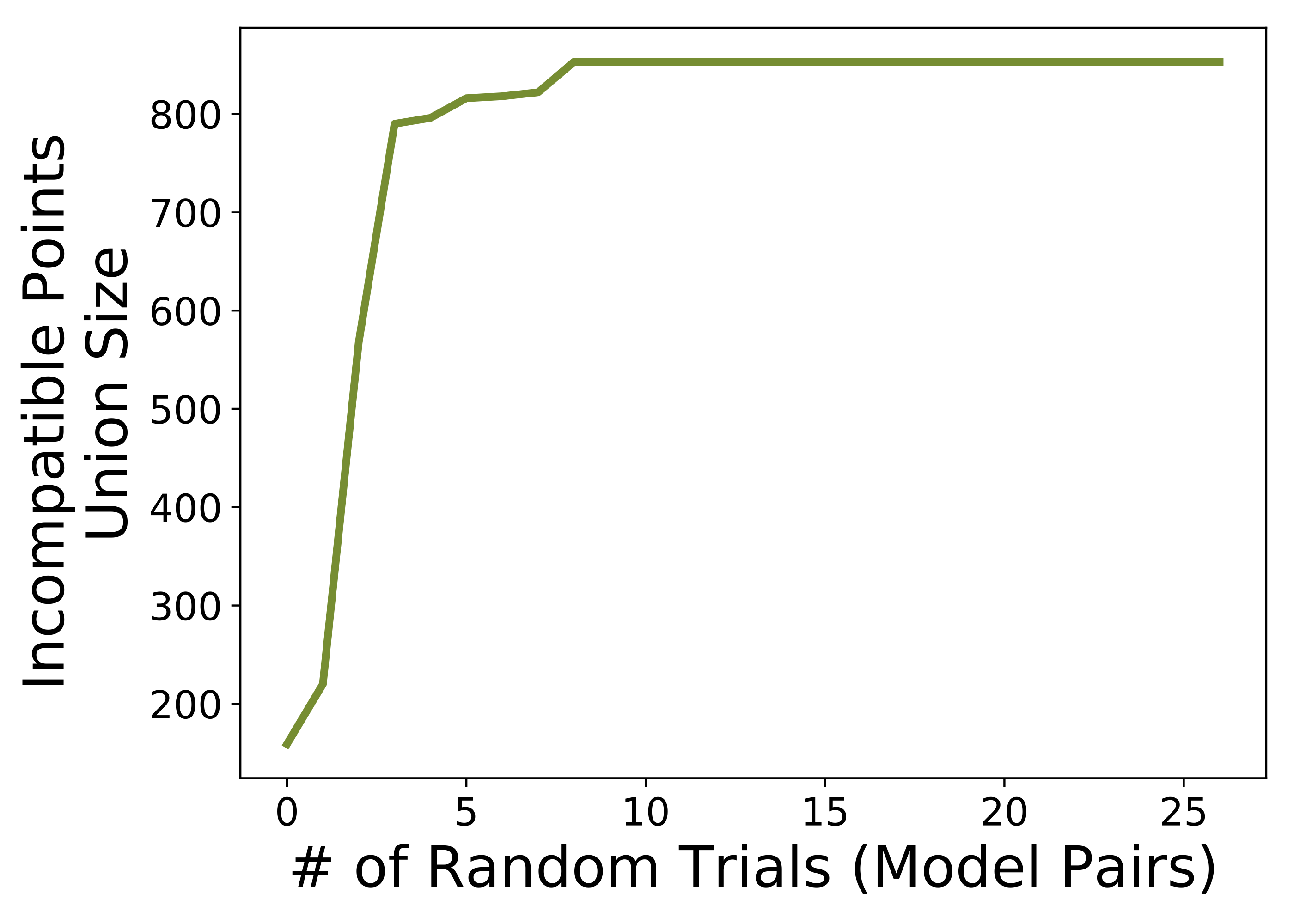}
    \caption{}
    \label{fig:saturation}
    \Description{}
    \end{subfigure}%
    \caption{(a) Distribution of incompatible examples over the output softmax score for two individual models. (b) Number of unique incompatible examples across pairs of models.}
\vspace{-4mm}
\end{figure}

As an upper-bound baseline on backward compatibility, we first seek to understand backward compatibility empirically when the \emph{training data has not been updated}. This means that the two training datasets, $D_1$ and $D_2$, are identical, and any discrepancies between $h_1$ and $h_2$ are purely due to optimization stochasticity. Possible sources of optimization stochasticity include random model weight initialization or  random data shuffling across training epochs. In this section specifically, model retraining  entails simply training a new model with a different random seed for random initialization and data shuffling.

To better understand backward compatibility in this simple baseline condition, we ask the following questions:
\begin{enumerate}
    \item Do the incompatible points correspond to datapoints for which $h_1$ and $h_2$ have low confidence in their predictions?
    \item Is there a correlation between incompatible points and example forgetting, as studied by \cite{toneva2018empirical}?
    \item Are the incompatible points between any two models the same across different trials? 
\end{enumerate}
Since deep networks trained with high-dimensional datasets are more subject to optimization stochasticity due to the non-convex nature of the optimization function in such settings, we first study this question on a Logistic Regression model on the FICO credit score binary classification task \cite{maas2011imdb}. In Section~\ref{section:exp} and~\ref{section:pipeline} we repeat the same experiments for larger models and data dimensionality and use these results as a baseline.

Across 25 trials (i.e., two logistic regression models), the average accuracy of both $h_1$ and $h_2$ (as there is no change in the training data) is \textbf{70.5} $\pm$ \textbf{0.86 (\%)}, with \textbf{BTC 94.5} $\pm$ \textbf{1.31 (\%)} and \textbf{BEC 78.3} $\pm$ \textbf{2.15 (\%)} backward compatibility scores. These results show that  backward compatibility concerns may arise even when the training data remains identical and the model architecture is simple.

\subsection{Confidence in predictions}
We initially hypothesized that, when the training data does not change, backward incompatibility arises from examples with low confidence that are close to the decision boundary, which might be subject to slight changes in the boundary.
Figure \ref{fig:confidence} shows a histogram of all points in $D_{\text{test}}$ grouped by the softmax output of two individual models $h_1$ and $h_2$, and demonstrates a trend towards low confidence (softmax output = $0.5$). However, there still exist points for which either $h_1$ or $h_2$ had high confidence, which is worrisome; a high prediction confidence by $h_1$ on certain points may cause a user or system to be especially reliant on that prediction.   

\subsection{Relationship with example forgetting}
\begin{table}[t]
  \caption{Relationship between incompatible examples and forgetting events on the validation set.}
  \footnotesize
  \centering
  \tabcolsep = 0.005\textwidth
  \label{tab:forg}
  \begin{tabular}{lll}
    \toprule
    Data Type & \# Forgetting & \ \# Forgetting  \\
            & Events ($h_1$ vs. $h_2$) & Events ($h_2$ vs. $h_1$) \\
    \midrule
    $h_1$ and $h_2$ both correct & $.51 \pm .12$ & $.56 \pm .16$\\
    $h_1$ and $h_2$ both incorrect & $.83 \pm .05$ & $.91 \pm .08$ \\
    $h_1$ correct, $h_2$ incorrect & $1.52 \pm .42$ & $1.77 \pm .2$\\
  \bottomrule
\end{tabular}
\vspace{-4mm}
\end{table}
Next, we measure the relationship between incompatible points and example forgetting, as studied by Toneva et. al~\citep{toneva2018empirical}. Because example forgetting measures the number of "forgotten events" by a model across epochs \textit{during training}, we use a separate validation set for this particular experiment to avoid touching the test set during training. More precisely, for each example in the validation set, we count the number of epochs when the model made an incorrect prediction (one "forgetting event") for a point it had previously predicted correctly. We then aggregate the average number of forgetting events per example by three different regions of interest as shown in Table~\ref{tab:forg}. This procedure is repeated for both models.

The observations suggest a strong correlation between the forgotten examples over the course of training and the incompatible examples between two models. Further experiments using deep neural networks demonstrate the same phenomena. However, while measuring example forgetting is costly and involves calculations at each training step, as noted in \cite{toneva2018empirical}, detecting similar points using backward compatibility only requires one pass at the end of training. Moreover, since using forgettable data points helps improve model learning (also noted in~\cite{toneva2018empirical}), our results suggest that simply considering backward compatibility when designing models can help identify points that would lead to similar performance boosts. Finally, there may also exist opportunities for combining both measurements to devise training algorithms that can ensure backward compatibility on-the-fly during training by guiding the model to remain compatible with earlier versions (i.e., checkpoints) of itself. 

\subsection{Consistency across multiple trials}
Finally, we ask whether incompatible points are a property of the data itself, or unique to different pairs of models $h_1$ and $h_2$. Figure \ref{fig:saturation} shows how the size of the union of all sets of incompatible test examples grows as the number of random trials (i.e., different pairs of  $h_1$ and $h_2$), increases. We observe a sharp increase between 0-5 pairs before a plateau, suggesting that, for this particular task, it is indeed possible that there exists a set of $~800$ datapoints (40.5\% of the test set) in which all incompatible points belong. However, as indicated by the sharp increase between 0 and 5 trials, the incompatible points between two models may be completely disjoint and even ensembling techniques, increasing model capacity with less than five models, may not alleviate the problem. 
  
\noindent \textbf{Results summary:}  Significant incompatibilities arise even when the training data remains identical and within a simple underlying model architecture. Importantly, our results show that backward compatibility captures a notion of "example difficulty", and, per the simplicity of identifying incompatible points, can be a useful tool for understanding the types of data that a model will be unreliable for when it comes to working with other components or end users.

\section{Backward Compatibility under Noisy Model Updates}
\label{section:exp}
We now study backward compatibility when additional data is added to training. We focus on noisy model updates where $D_2$ is a larger, noisier dataset than $D_1$. In all of the following experiments, $h_2$ is \textit{retrained} on $D_2$ and initialized by using $h_1$'s weights, as we found that doing so consistently improved both model accuracy and backward compatibility. Yet, our results still show that, despite strong overall accuracy gains, backward incompatibility still persists and is further exacerbated as dataset noise increases.

In the following experiments, we consider noise biased by class (MNIST digit and CIFAR10 image category) as well as noise on class-independent groups (IMDB movie review genre). We specifically consider three types of noise:
\begin{figure*}[ht]
\centering
\begin{subfigure}[t]{0.25\linewidth}
  \centering
  \includegraphics[width=\linewidth]{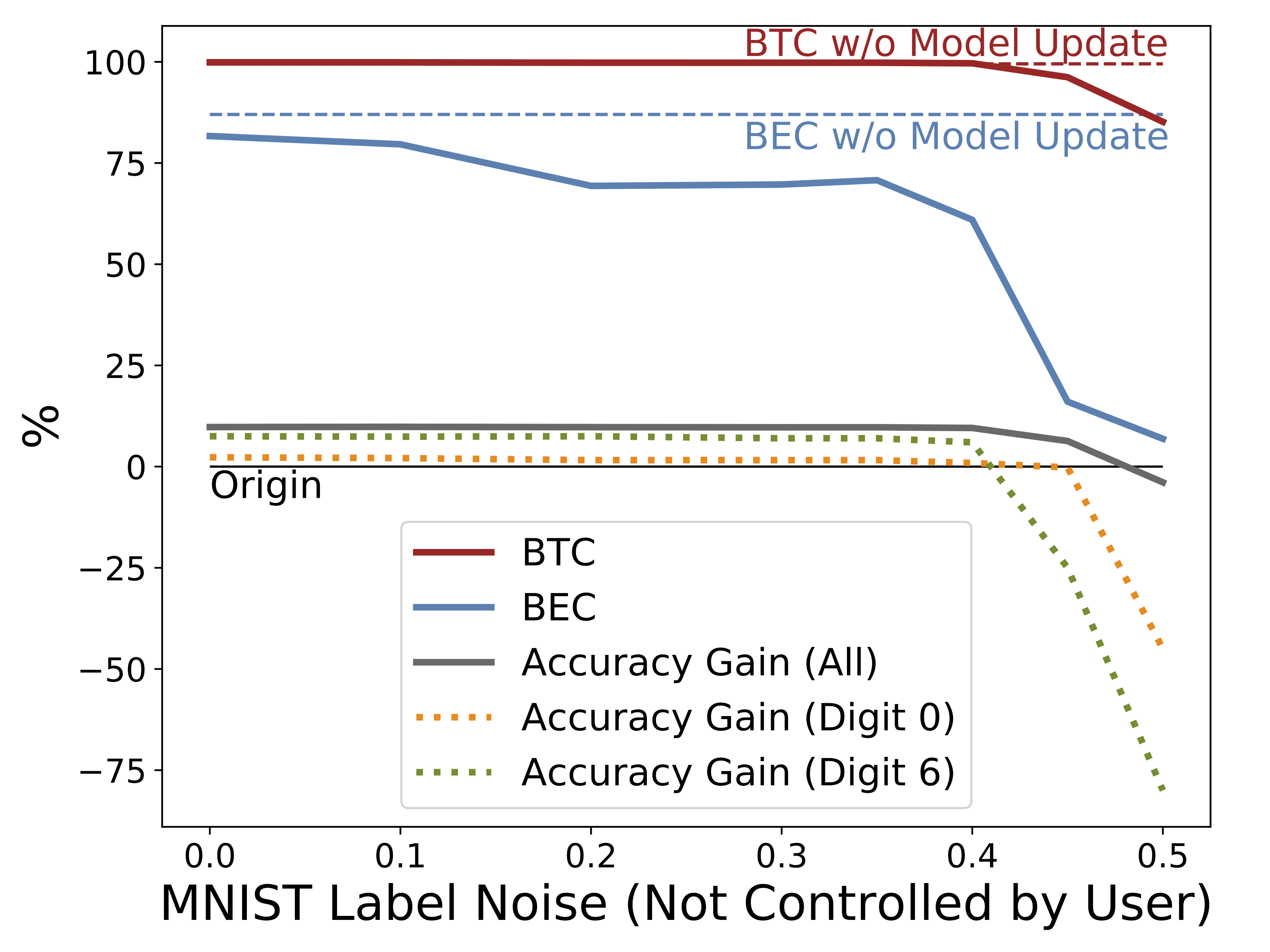}
  \caption{MNIST label noise.}
  \label{fig:sub1}
\end{subfigure}%
~\hspace{-5pt}
\begin{subfigure}[t]{0.25\linewidth}
  \centering
  \includegraphics[width=\linewidth]{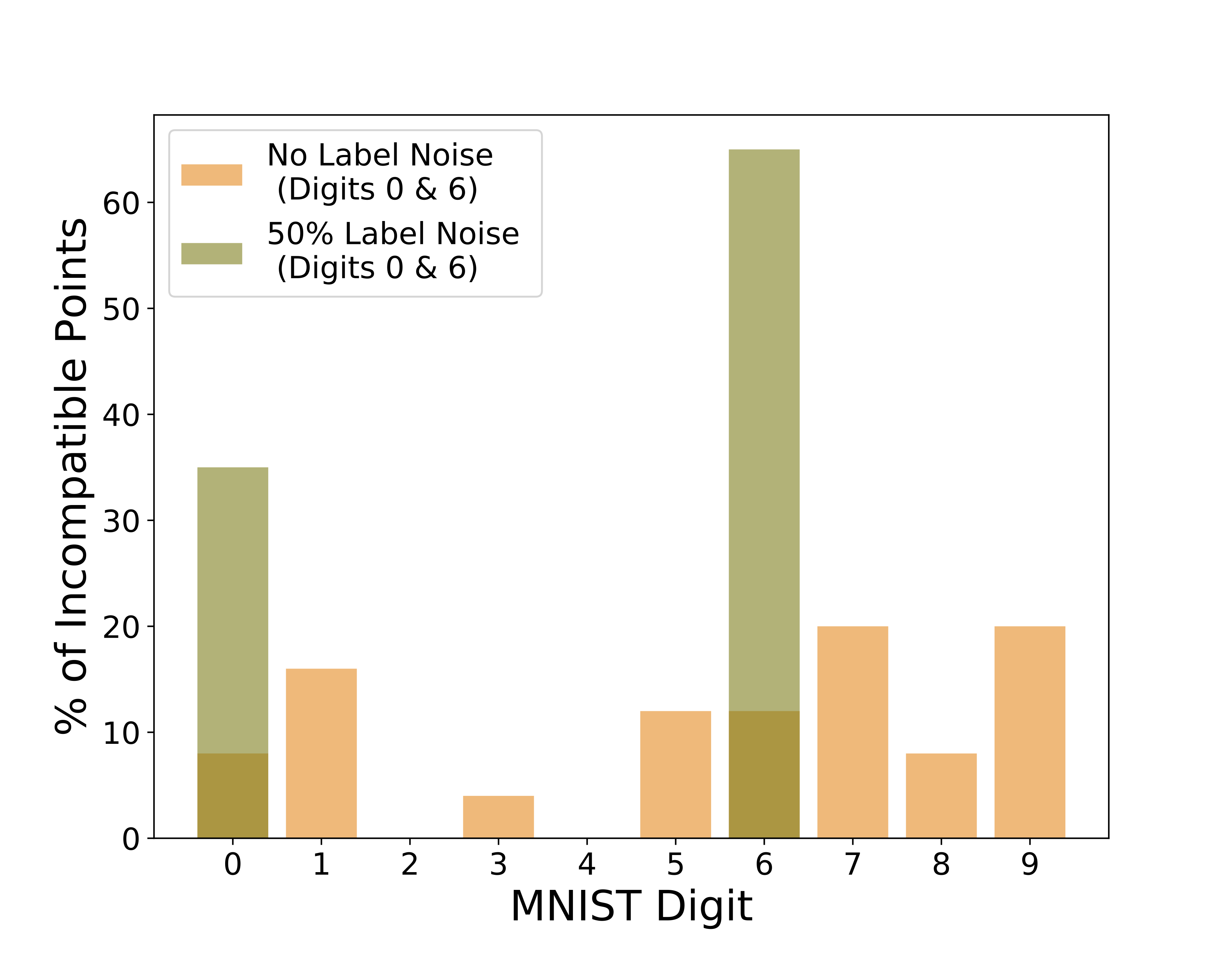}
  \caption{MNIST label noise.}
  \label{fig:sub2}
\end{subfigure}%
~\hspace{-5pt}
\begin{subfigure}[t]{0.25\linewidth}
  \centering
  \includegraphics[width=\linewidth]{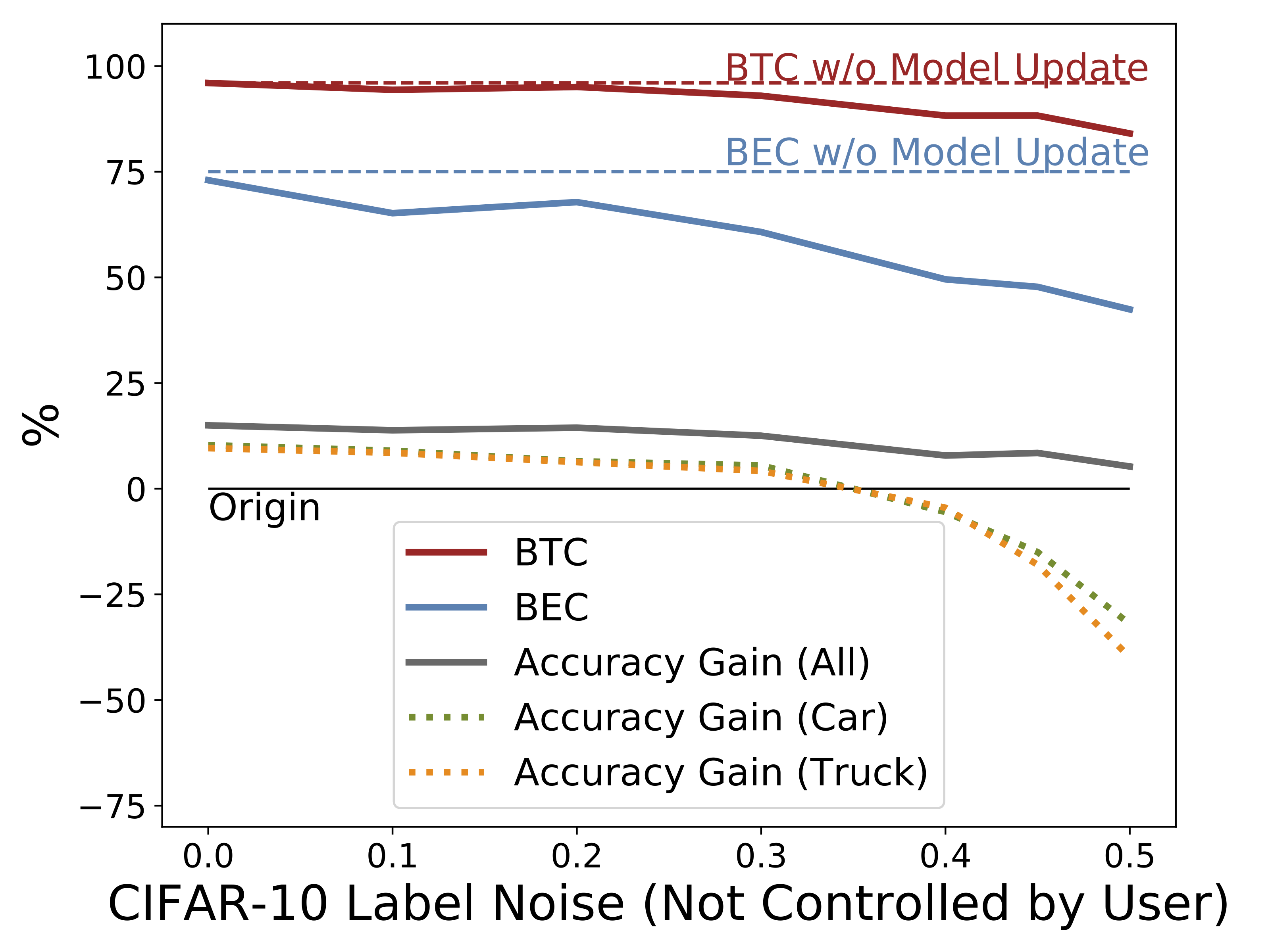}
  \caption{CIFAR-10 label noise.}
  \label{fig:sub3}
\end{subfigure}%
~\hspace{-5pt}
\begin{subfigure}[t]{0.25\linewidth}
  \centering
  \includegraphics[width=\linewidth]{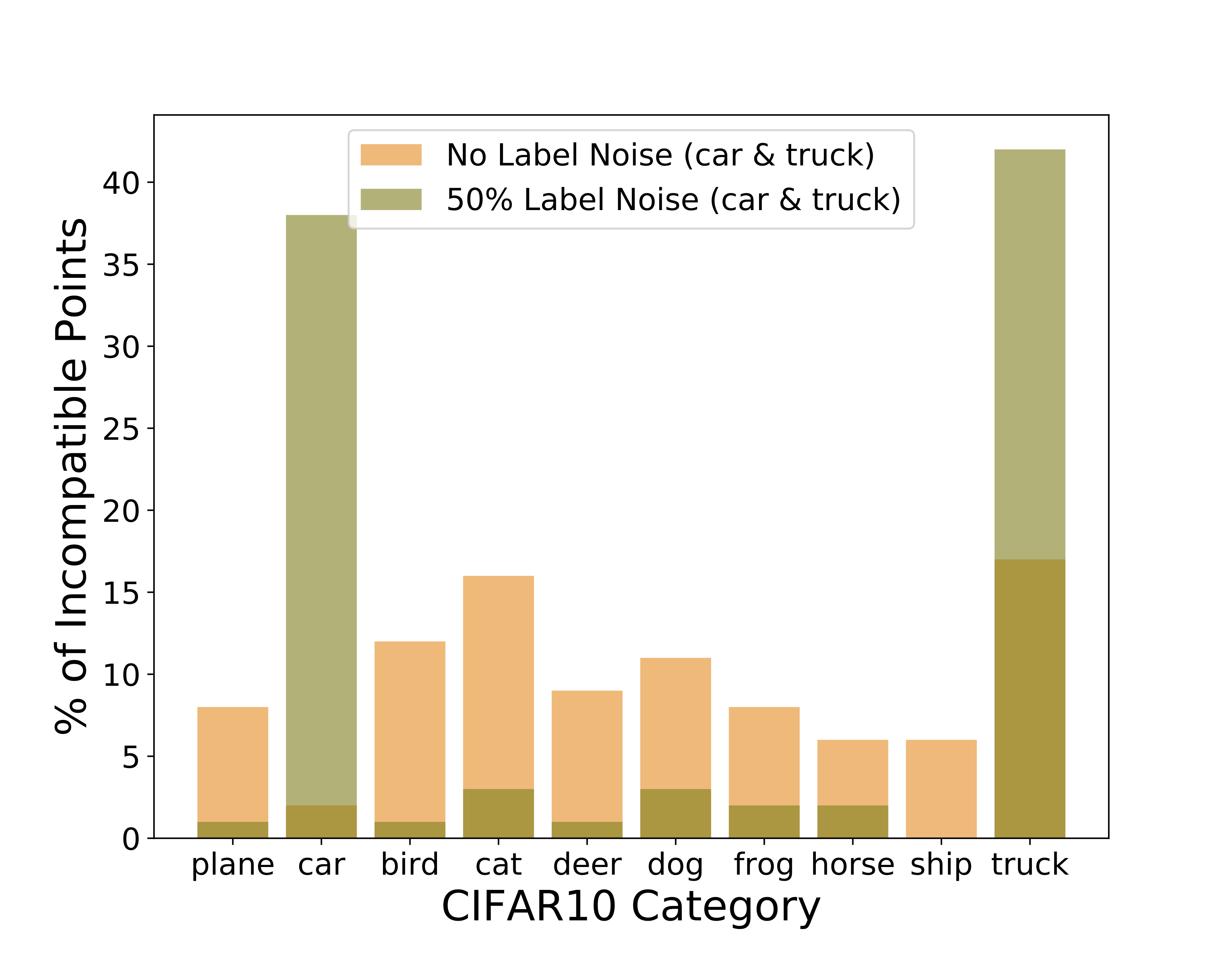}
  \caption{CIFAR-10 label noise.}
  \label{fig:sub4}
\end{subfigure}%

\begin{subfigure}[t]{0.25\linewidth}
  \centering
  \includegraphics[width=\linewidth]{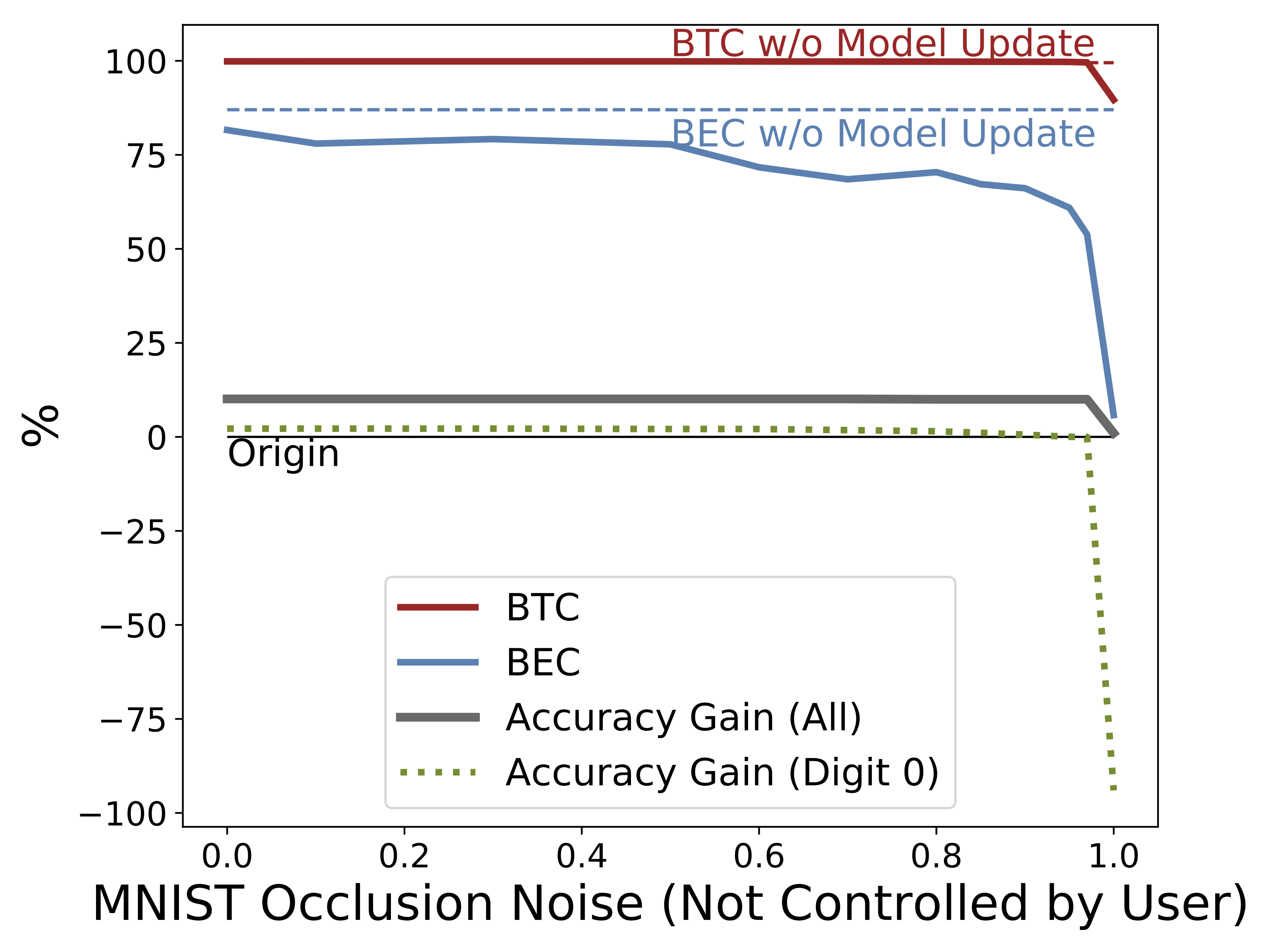}
  \caption{MNIST feature noise.}
  \label{fig:sub5}
\end{subfigure}%
~\hspace{-5pt}
\begin{subfigure}[t]{0.25\linewidth}
  \centering
  \includegraphics[width=\linewidth]{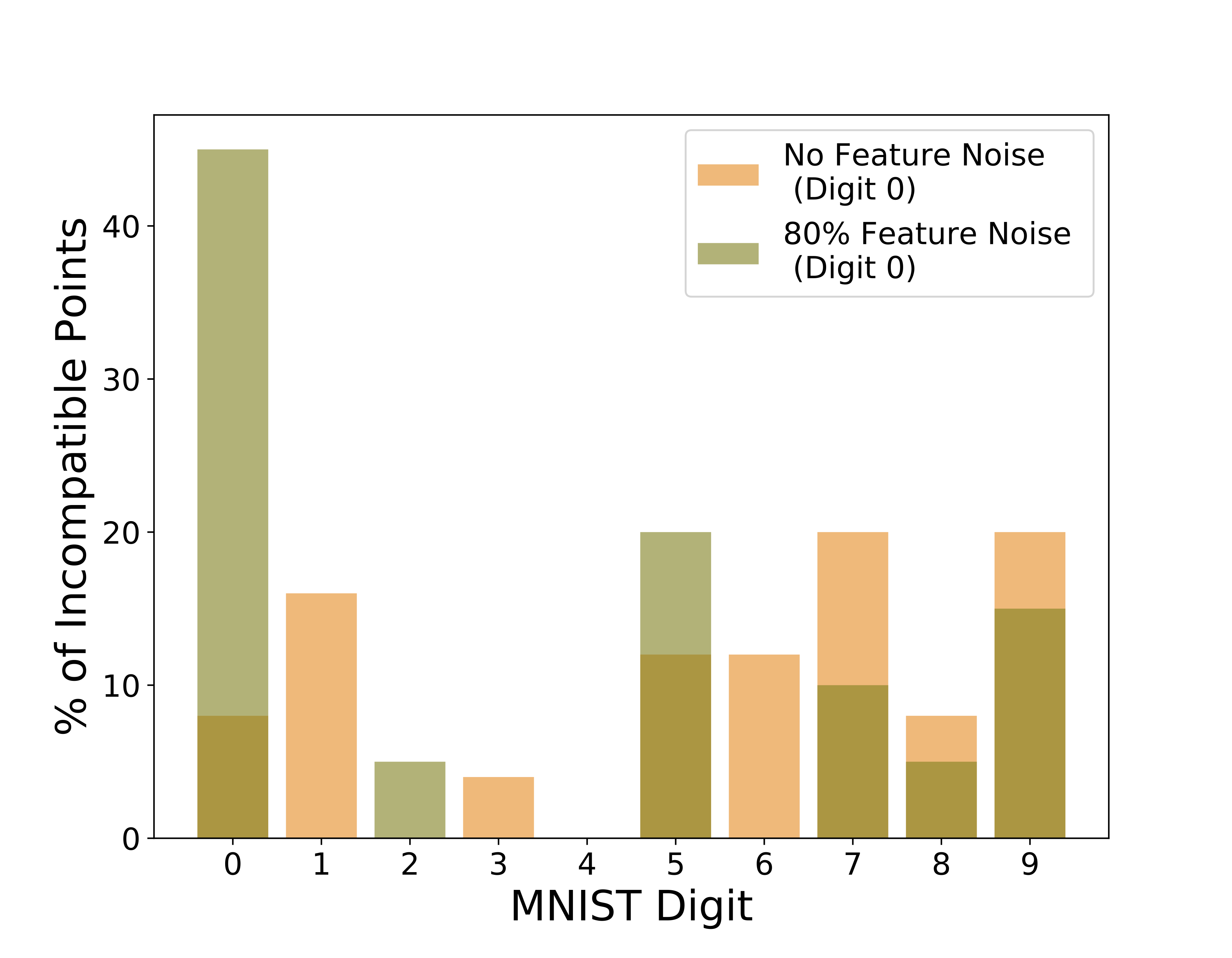}
  \caption{MNIST feature noise.}
  \label{fig:sub6}
\end{subfigure}%
~\hspace{-5pt}
\begin{subfigure}[t]{0.25\linewidth}
  \centering
  \includegraphics[width=\linewidth]{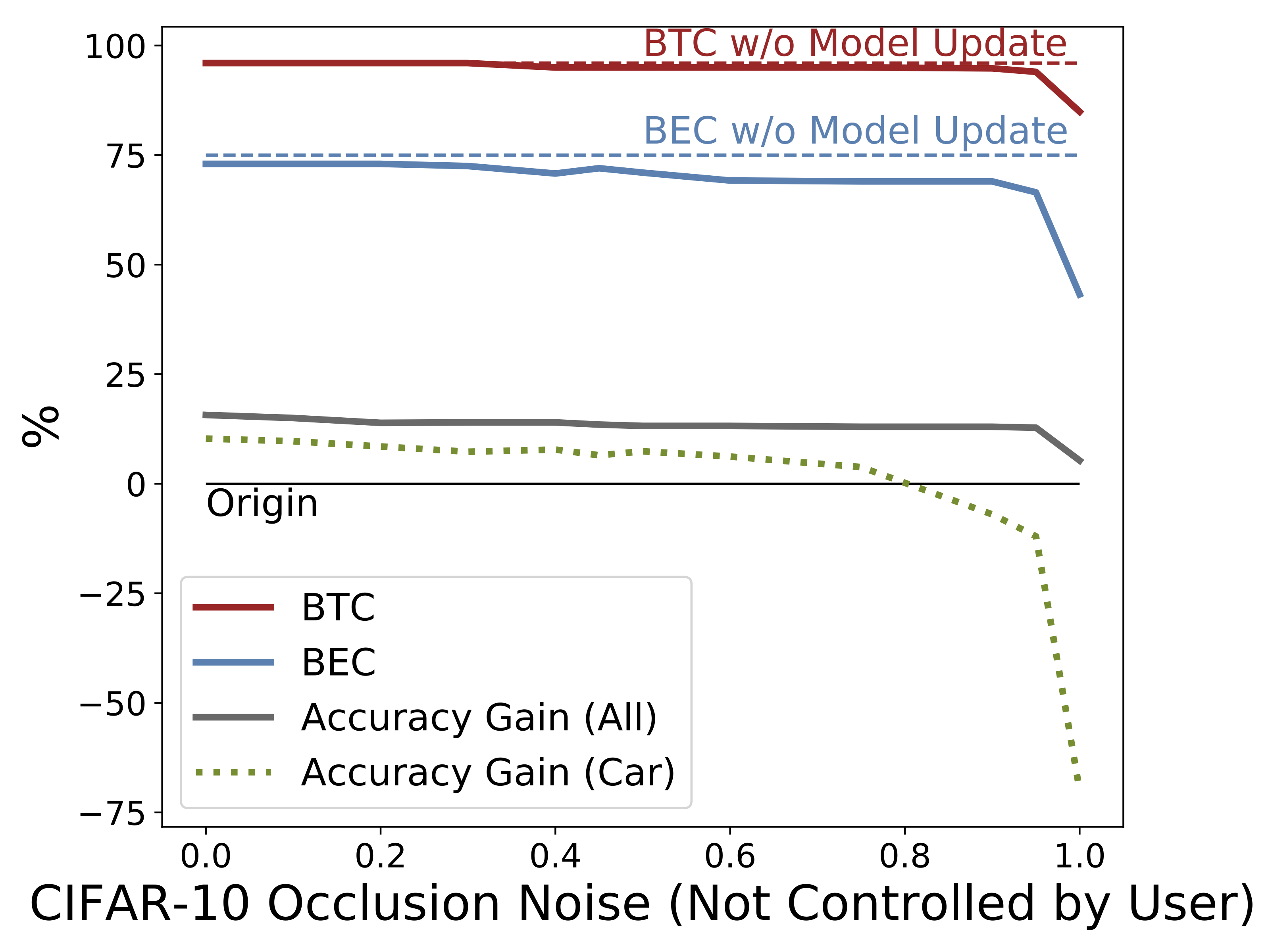}
  \caption{CIFAR-10 feature noise.}
  \label{fig:sub7}
\end{subfigure}%
~\hspace{-5pt}
\begin{subfigure}[t]{0.25\linewidth}
  \centering
  \includegraphics[width=\linewidth]{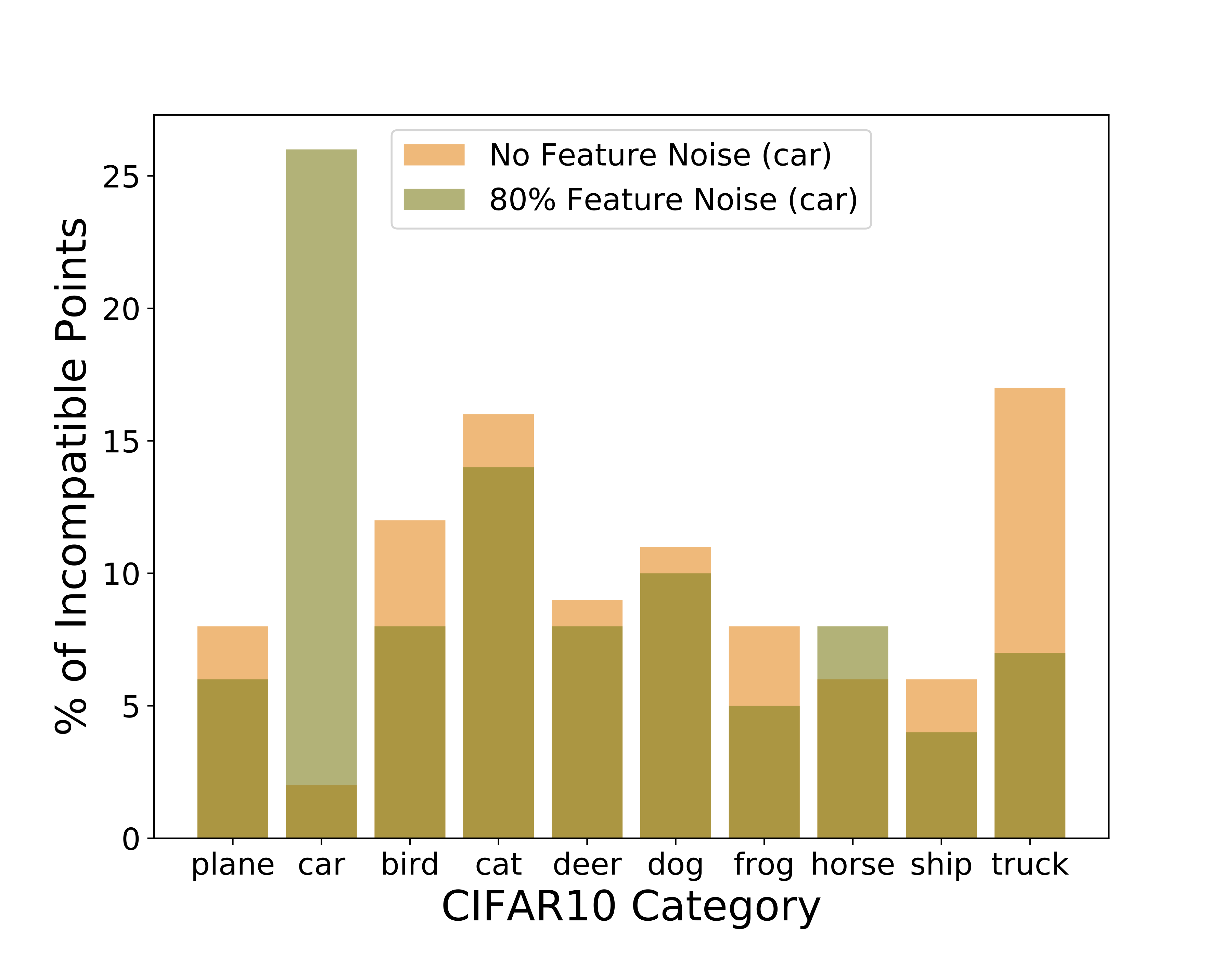}
  \caption{CIFAR-10 feature noise.}
  \label{fig:sub8}
\end{subfigure}

\begin{subfigure}[t]{0.25\linewidth}
  \centering
  \includegraphics[width=\linewidth]{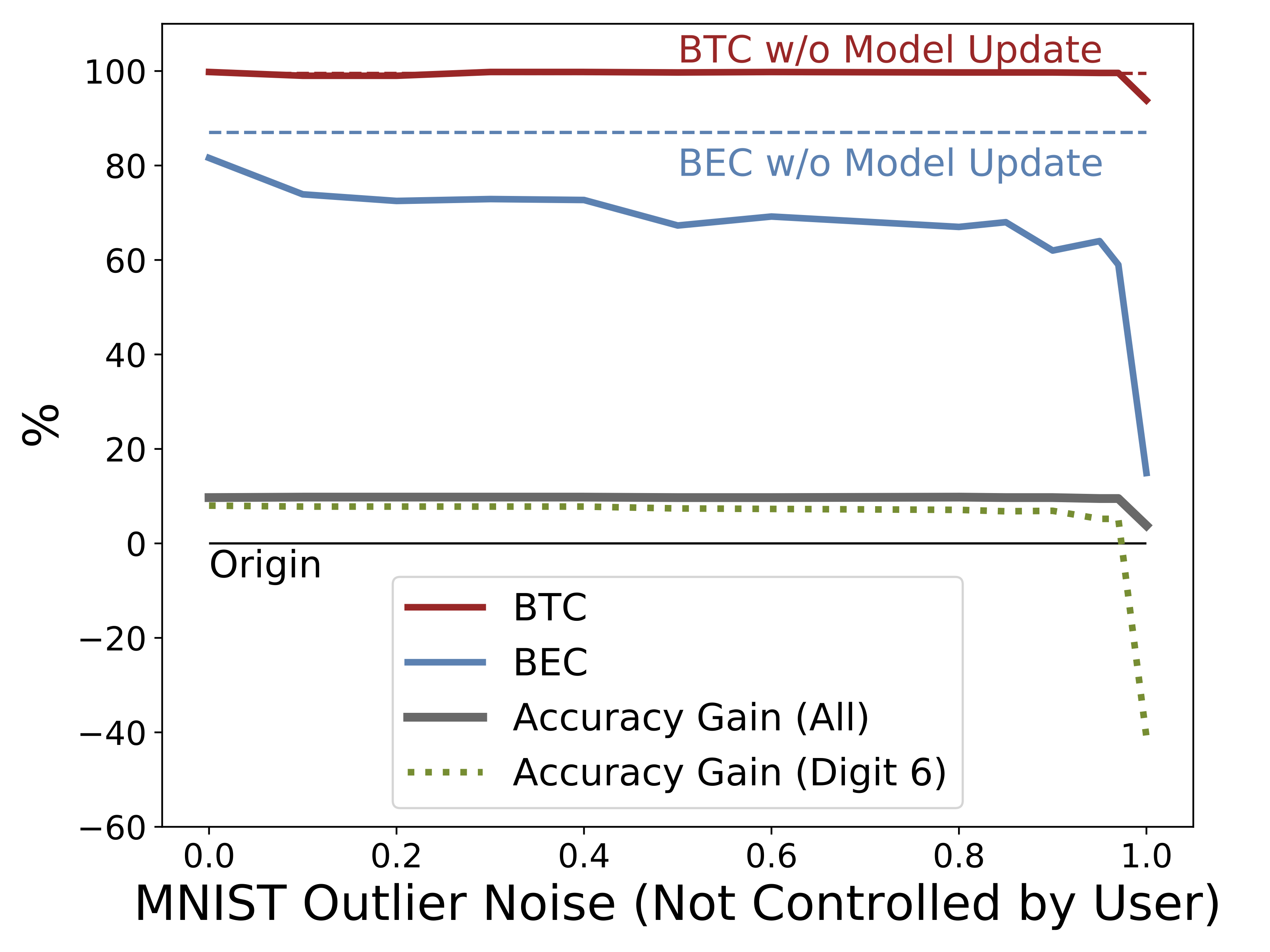}
  \caption{MNIST outlier noise.}
  \label{fig:sub9}
\end{subfigure}%
~\hspace{-5pt}
\begin{subfigure}[t]{0.25\linewidth}
  \centering
  \includegraphics[width=\linewidth]{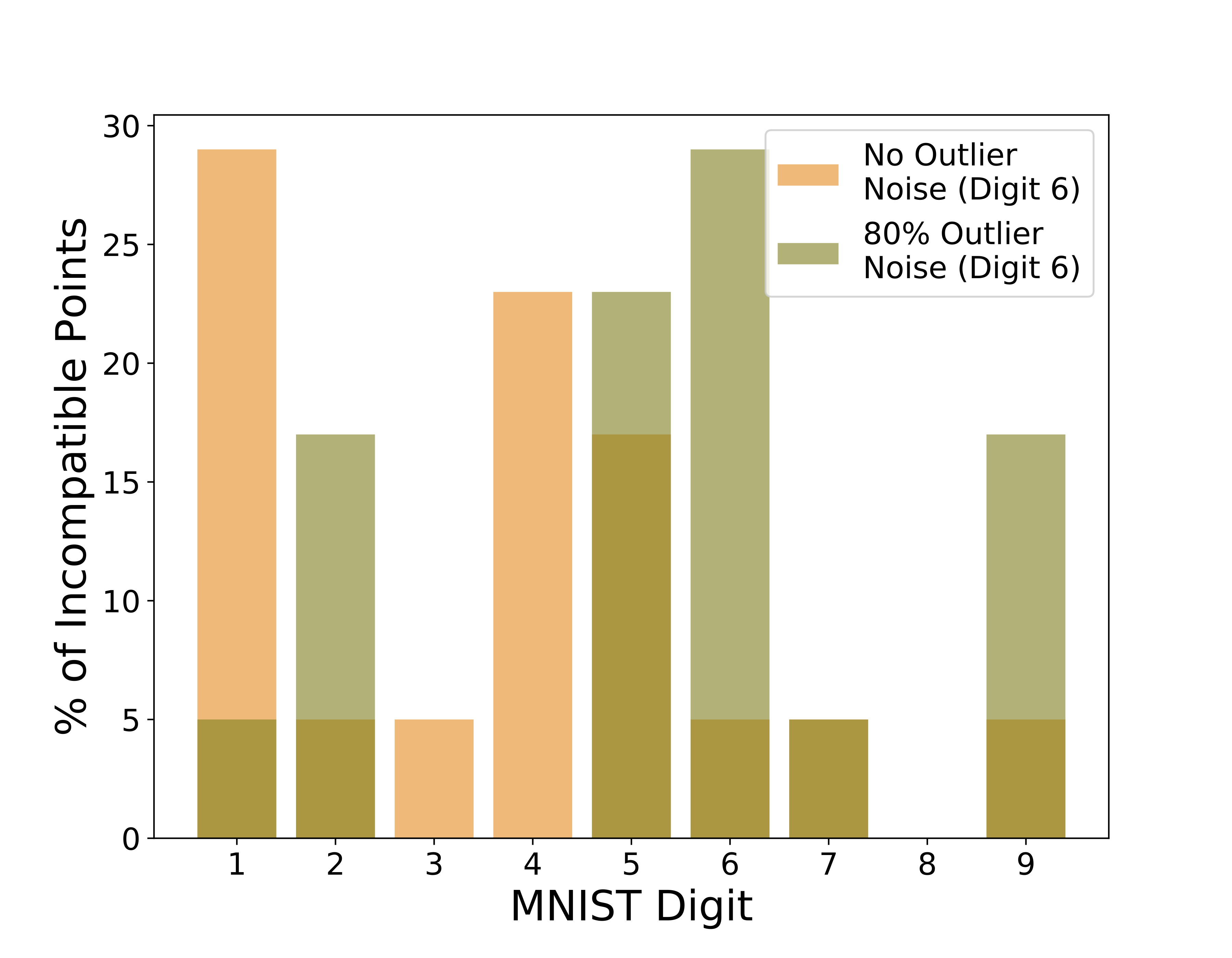}
  \caption{MNIST outlier noise.}
  \label{fig:sub10}
\end{subfigure}%
~\hspace{-5pt}
\begin{subfigure}[t]{0.25\linewidth}
  \centering
  \includegraphics[width=\linewidth]{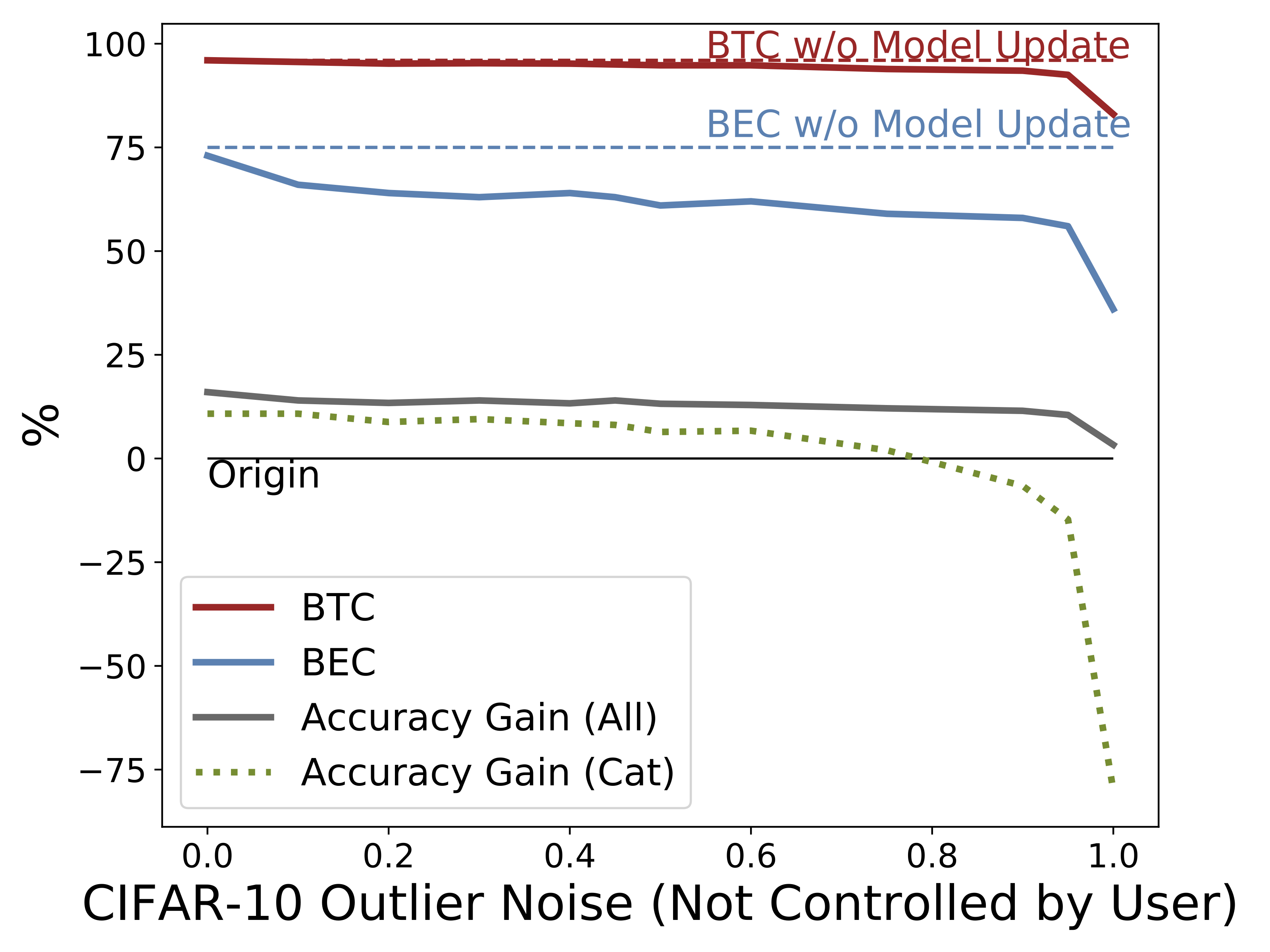}
  \caption{CIFAR-10 outlier noise.}
  \label{fig:sub11}
\end{subfigure}%
~\hspace{-5pt}
\begin{subfigure}[t]{0.25\linewidth}
  \centering
  \includegraphics[width=\linewidth]{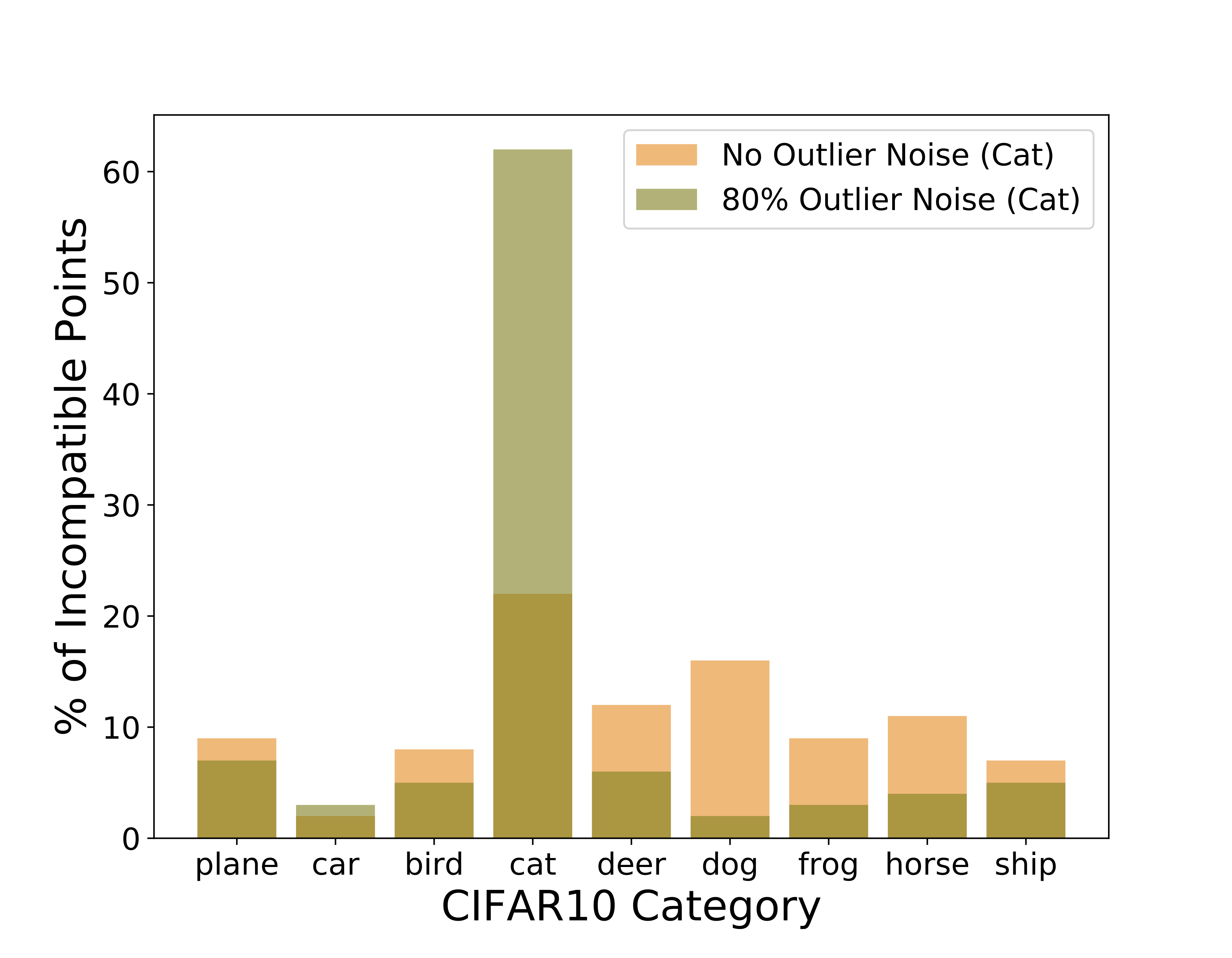}
  \caption{CIFAR-10 outlier noise.}
  \label{fig:sub12}
\end{subfigure}
\caption{Backward compatibility with varying noise for MNIST and CIFAR-10.}
\label{fig:test}
\vspace{-4mm}
\end{figure*}

\begin{enumerate}
    \item \emph{Label noise} - Instances from one class are labeled in error as another similar class (e.g., digit "0" as letter "o" in OCR). 
    \item \emph{Feature noise} - Features of the input instances themselves are noisy (e.g., occlusion for images, noisy sensors etc.).
    \item \emph{Outlier noise} - Instances from a class not part of the task definition are wrongly labeled with a class that is part of the task definition (e.g., lion images may be added to the CIFAR dataset and be classified as cats). 
\end{enumerate}

\subsection{Class-level incompatibility}
We first analyze backward compatibility in model updates where the noise is biased towards a particular set of target \textit{classes}.

\subsubsection{Label Noise}
We simulate label noise, where instances from one class are labeled erroneously as a similar class, using the digit pair (0,6) and CIFAR-10 category pair ("car", "truck") as the target label pairs that are switched. Such label noise can occur through annotation noise, for example, where a human annotator via crowdsourcing may accidentally label an image of a truck with "car," or a painted digit 6 on a street address as "0" due to the similar shape. The degree of noise affects the likelihood of switching the labels of datapoints belonging to the target label pairs. In the no-noise condition on a held-out test set, the three-layer CNN achieves $99.4\%$ accuracy after 50 epochs for MNIST, while for CIFAR-10 the ResNet-18 model achieves $90.3\%$ overall accuracy after 35 epochs.

Figures \ref{fig:sub1} and \ref{fig:sub3} demonstrate the effect of varying label noise on BTC (solid red), BEC (solid blue), gain in overall accuracy by $h_2$ over $h_1$ (solid gray), and gain in class-level accuracies by $h_2$ over $h_1$ (dashed green and orange). The dashed red and blue lines show BTC and BEC baselines when there is no model update, and $D_1 = D_2$. For both datasets, we observe a decrease in backward compatibility even when \textit{the accuracy gain is positive}. For example, for 30\% label noise, despite $h_2$ improving overall accuracy and class-level accuracies for the target labels, there is a notable decrease in BEC, as well as a minor decrease in BTC. Interestingly, we notice that the BTC measure decreases more for CIFAR-10 than MNIST, likely due to the extremely high accuracy of $h_2$ on MNIST. Most importantly, for higher noise values ($>0.4$), the target class accuracy starts to drop even though the overall accuracy gain stays positive. 

Further analysis of incompatible points (see histograms in Figures \ref{fig:sub2} and \ref{fig:sub4}) shows that, as label noise increases to 50\%, the proportion of incompatible points within noise susceptible classes increases. These results demonstrate that backward compatibility analyses can help identify the types of data points that were subject to biased noise at the time of data collection. 
\vspace{-2pt}
\subsubsection{Data Feature Noise}

We next simulate feature noise, focusing on \textit{occlusion}, where images are blocked by artifacts such as smudges and out-of-domain objects. We randomly distribute occlusions that occupy $\sim$20\% of target images (Digit 0 in MNIST, "Car" in CIFAR-10). Real-world examples of occlusion include artifacts from the camera or current context (such as rainy weather) blocking a subject.

Interestingly, Figure \ref{fig:sub5} shows that, while for MNIST the accuracy gain on the target noise class (Digit 0) is stable up until high amounts of noise, $\text{BEC}$ decreases early on. This may be due to the simplicity of the MNIST task and retraining with $h_1$'s weights, which may have already learned the digit $0$ well (explaining the minimal gains on Digit 0 by $h_2$). Adding noise may first decrease performance on specific points before hurting the entire class representation itself. However, for CIFAR-10, because $h_2$ strongly improves the "Car" class accuracy, small amounts of noise may harm performance only on parts of the data that $h_1$ anyways performed poorly on, resulting in a small backward compatibility decrease. Finally, Figures \ref{fig:sub6} and \ref{fig:sub8} show that at 80\% noise, analysing the incompatible points can identify the target groups of the biased noise.  

We note that adding feature noise sometimes \textit{improves} performance by reducing overfitting \cite{holmstrom1992noise}, and is thus used for data augmentation. This may explain why backward compatibility significantly decreases only with a high degree of occlusion noise ($> 80\%$).

\subsubsection{Outlier Noise}
To simulate outlier noise, we include instances in the training data that are not part of the initial task, but were labeled as one of the in-task classes. We convert both the MNIST and CIFAR-10  tasks to predictions over nine classes, treat Digit 0 and "Truck" as outliers, and  Digit 6 and "Cat" as targets for the outlier noise. Semantically, this means that the training data will contain images of trucks (i.e., Cat$\circledR$  trucks by the Caterpillar series) that are labeled as "cat" because of language ambiguity.

While the main backward compatibility trends  persist here as well, Figures \ref{fig:sub10} and \ref{fig:sub12} highlight that it is possible that noise in the targeted class may decrease compatibility in other classes as well. For instance, the number of incompatible points on MNIST increases on digits "2", "5", "6", "9" even though only digit "6" was directly affected by noise. Such behavior can be extenuated when the classification task contains more classes and the outlier examples look similar to many of the classes represented in the task.

\noindent \textbf{Results summary:} Adding noise biased towards a class is aligned with lack of backward compatibility for that class. However, the effects can propagate to increased incompatibility of other classes, especially if these classes become less distinguishable after the inclusion of outliers. Measuring backward compatibility helps detect this unreliability even when the overall accuracy gain by $h_2$ is high.

\begin{figure}[t]
    \centering
    \includegraphics[width=0.37\textwidth]{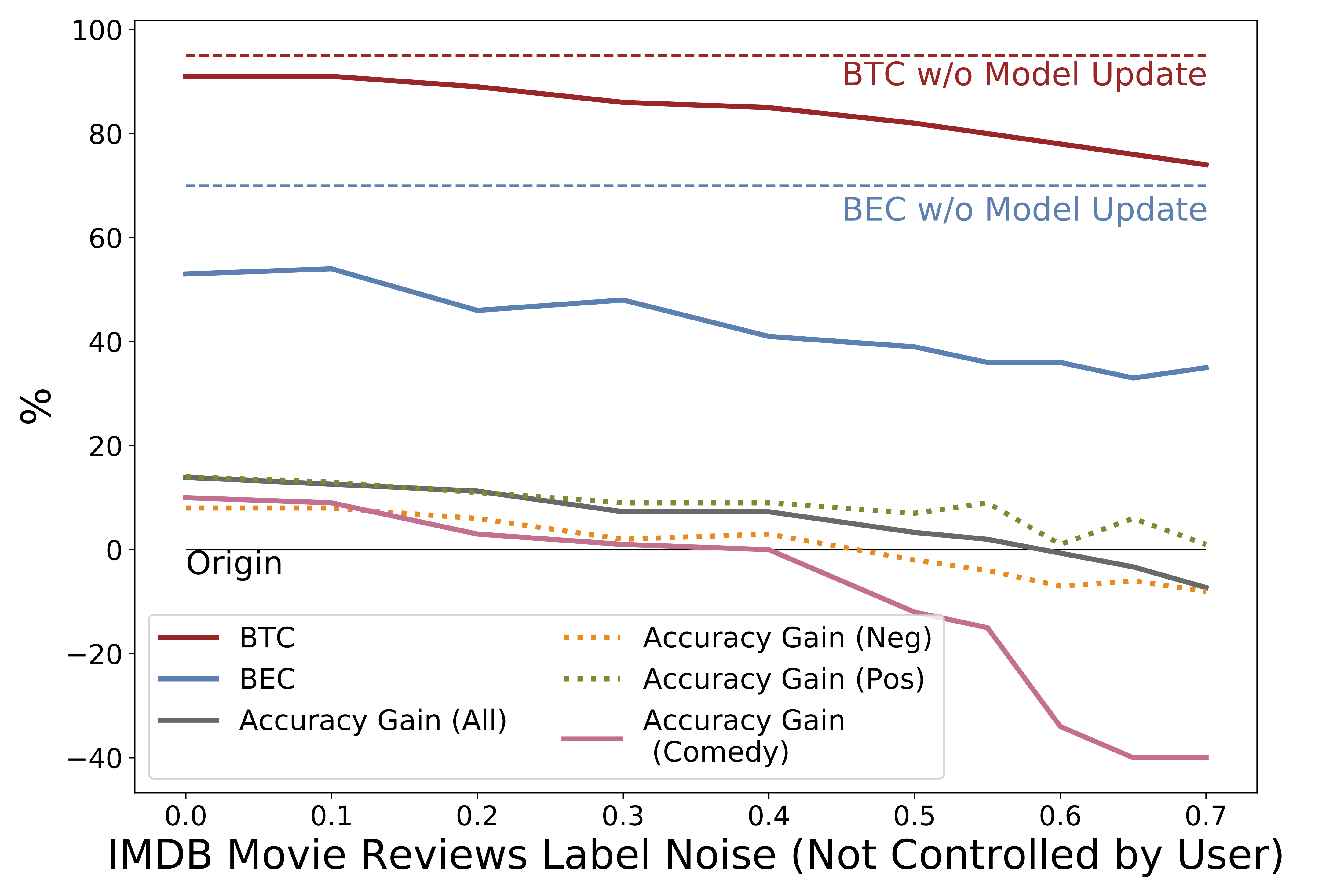}
    \caption{Backward compatibility analysis on an IMDB movie reviews sentiment classification task.}
    \label{fig:imdb}
\end{figure}
\subsection{Beyond class groups: Sentiment Analysis}
A natural question following the previous results is: \emph{Can backward incompatibility be identified by only monitoring the accuracy of individual classes?} Here we argue that this approach, although complementary, is unsatisfying because: 
\begin{enumerate}
    \item In the previous experiments, there still existed a range where $h_2$ \textit{increased accuracy} for a class yet backward compatibility \textit{decreased}, likely due to performance decrease for a \textit{sub-class}.
    \item A possible grouping of datapoints that do not share the same class label may suffer from backward compatibility which a per-class analysis would hide.

\end{enumerate}In essence, measuring backward compatibility accounts for all types of points for which $h_2$ performance decreases, without needing to define and analyze a priori which points to examine. 

To demonstrate this, we analyze backward compatibility in a movie review sentiment analysis task considering data noise biased on groups beyond labels - specifically, all movies belonging to a specific \textit{genre}. Consider a large dataset of movie reviews created by scraping reviews from different web pages, including genre-specific web pages such as a comedy review site or sub forum. The labels may be scraped from star ratings accompanying the reviews. If the comedy site changes its html structure (which happens frequently in websites), the scraper may fail to map the review to the right rating and therefore induce biased noise for the comedy genre.

We simulate this setting using the IMDB Movie Reviews dataset \cite{maas2011imdb}. We filter all reviews in the training data with the keyword "comedy," and then flip the label with varying noise. Figure \ref{fig:imdb} summarizes the results, including the comedy group accuracy gain (solid pink). The test performance drop for comedy reviews is far more severe than the performance drop for either negative or positive classes. Furthermore, for noise between $0.4$ and $0.47$, there exists a region where \textit{class-based accuracy improves} on both classes but \textit{sub-class comedy group accuracy decreases}. An analysis of the incompatible points at $0.47$ noise, of which $59.5\%$ are comedy (vs. $20\%$ at 0 noise), would help detect this failure while class-based accuracy metrics would not.

\noindent \textbf{Results summary:} When backward incompatibility affects groups of data but not a whole class, monitoring class accuracy is insufficient. Moreover, since there may exist many possible groups, and it is not possible to know a priori which groups suffer from noise, the decrease in BEC and BTC scores and analyzing the incompatible points may be the only way to detect partial model regress. 
\begin{figure}[t]
    \centering
    \includegraphics[width=0.33\textwidth]{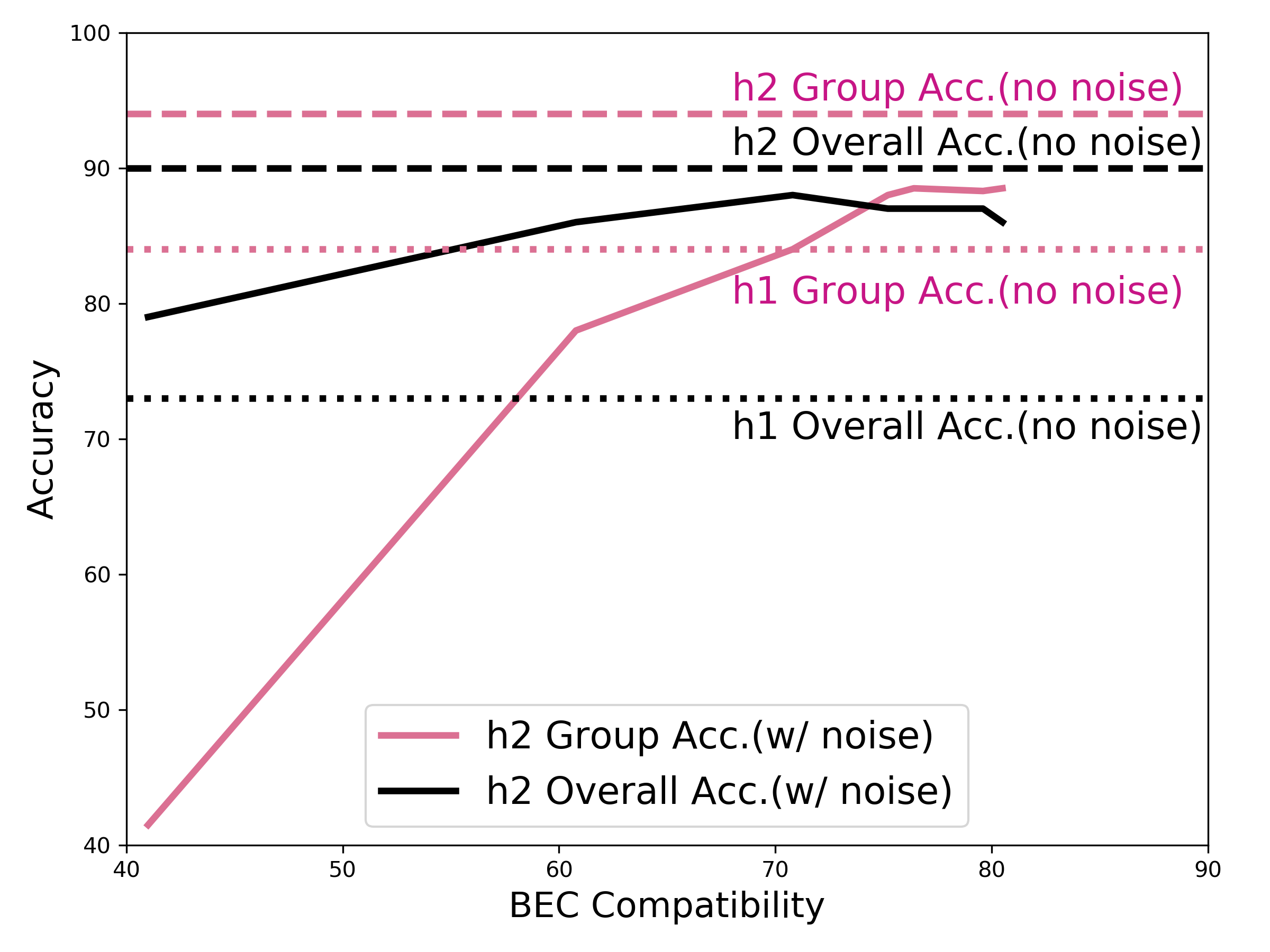}
    \caption{Regularization improves compatibility, but it may not fully prevent discrepancies from biased noise.}
    \label{fig:penalty}
\end{figure}
  \subsection{Effect of Regularization}
 We now highlight the benefits and limitations of regularization, introduced by prior work in the context of backward compatibility, under the presence of noisy model updates ~\cite{bansal2019updates}. The work proposes a new loss function to penalize newly introduced errors, enabling engineers to adjust the accuracy versus compatibility tradeoff during retraining. The loss penalizes the new model $h_2$ by its classification loss $L$ whenever the initial model $h_1$ is correct.  
 \vspace{-1mm}
 \begin{equation}
\begin{split}
    \text{ $L_c$} = \text{$L+\lambda_c\cdot (\mathds{1}(h_1(x)=y)\cdot L)$}  
    \end{split}
 \vspace{-2mm}
\end{equation}

Increasing the penalty $\lambda_c$ increases the compatibility of $h_2$ with respect to $h_1$. Using the same settings on the CIFAR-10 dataset under label noise as in Section 6.1.1, we vary $\lambda_c$ and measure compatibility, overall accuracy, and subgroup accuracy on the categories subject to noise (cars and trucks). Figure \ref{fig:penalty} shows that while regularization improves both accuracy and compatibility of $h_2$, the subgroup accuracy is often worse than the overall accuracy, and also unable to reach the optimal accuracy of $h_2$ without any noise. Given the high initial performance of $h_1$ on the subgroup, this observed limitation suggests that future techniques especially designed for backward compatibility \emph{under noise} can be more effective in obtaining the biggest gains from model updates.

\noindent \textbf{Results summary:} Prior methods to address backward compatibility help to improve performance under noisy updates to some extent, but there exists a need for stronger techniques that are aware of the source of incompatibility (e.g., biased noise).   
  \begin{figure}
        \centering
        \includegraphics[width=0.7\linewidth]{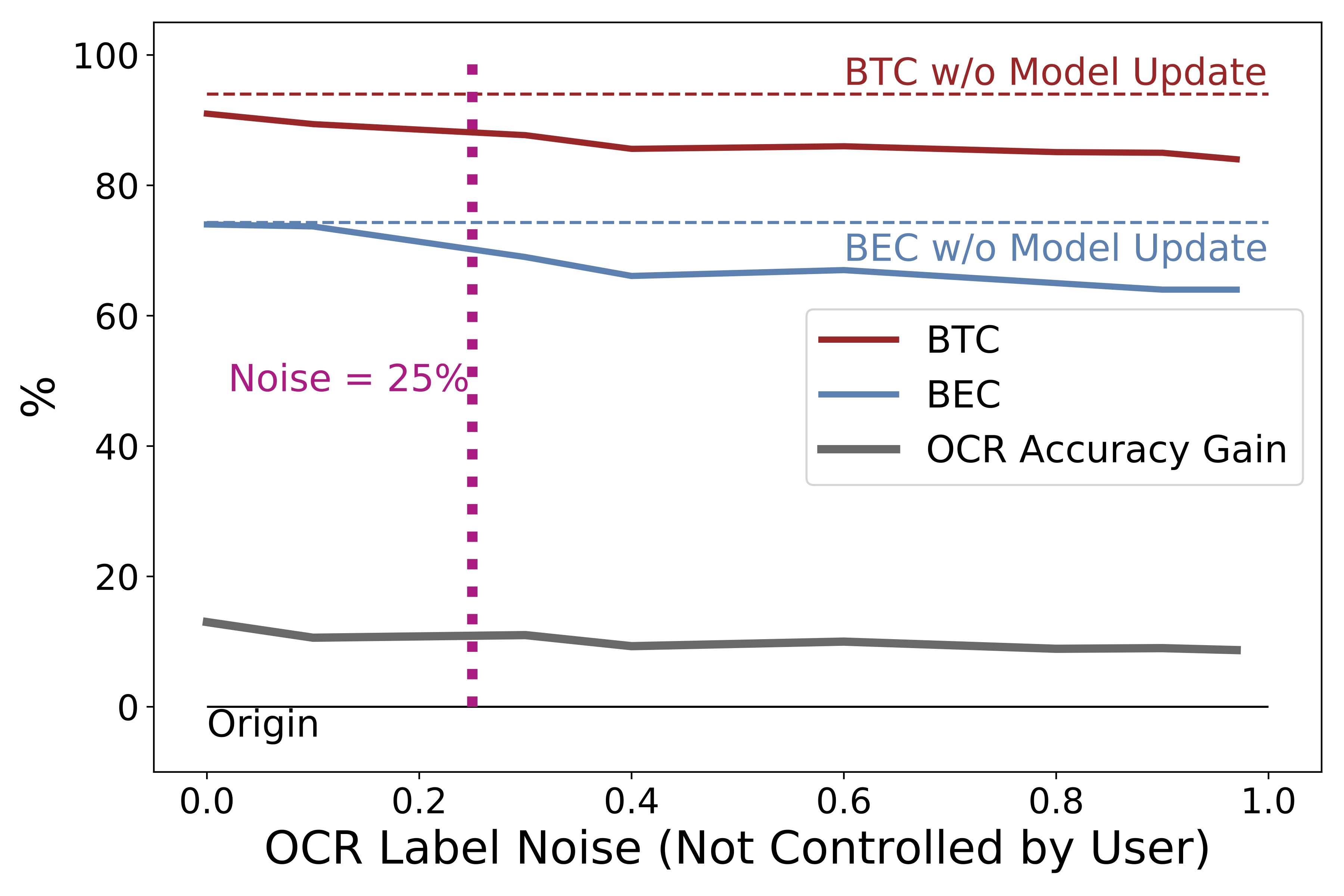}
        \caption{Backward compatibility analyses for the OCR pipeline component.}
        \label{fig:bc_ocr}
  \end{figure}

\section{Backward Compatibility in Machine Learning Pipelines}
\label{section:pipeline}
Finally, we position the analysis from the previous experiments in the context of the ML pipeline described in Section \ref{section:intro}, and demonstrate  how not controlling for backward compatibility during updates can lead to downstream degradation for specific input spaces. 

Recall our example of an off-the-shelf OCR model being used for downstream tasks by non-experts, such as receipt fraud detection. Consider the case where, in an effort to improve the OCR model's performance on a variety of fonts, the model is updated with a large, noisy dataset created from scraping the internet for images of different characters and asking people to label them through CAPTCHA tasks~\cite{von2003captcha}. Common mistakes by human annotators, such as mixing up "i" and "l", or "o" and "0" because of occlusion or rushing through the task, may occur often leading to decreased performance on these inputs. A fraud detection system that relies on the classifier's ability to discriminate between "i" and "l" or "o" and "0", will suddenly experience unexpected failures by misrecognizing spoofed terms such as "N1ke" for legitimate companies (i.e., "Nike"). 
  
We simulate this setting by assuming a text localization model using bounding boxes to detect characters followed by a character recognition model  trained on the Char74k dataset~\cite{deCampos09}. As in the previous experiments, we train $h_1$ on a clean, smaller subset of this data (20\%), and then re-train $h_2$ using the entire training data with varying degrees of label noise (swapping "i" w/ "l" and "o" w/ "0"). Figure \ref{fig:bc_ocr} shows the backward compatibility analyses for this update. 

Now, let us assume that the large dataset collection resulted in $25\%$ label noise, but gives $h_2$ a strong $14\%$ improvement over $h_1$.  The \textbf{89\% BTC} and \textbf{71\% BEC} scores suggest that there has been a decrease in backward compatibility when compared to the baselines. A quick glance through the incompatible points reveals examples such as those in Table~\ref{tab:ocr_examples}, the most common classes, with digit "0" and lower case letter "l"  consisting of 19\% and 16\% of all incompatible points, respectively. Interestingly, the upper case "Z" often appears in the set of incompatible points despite noise not directly influencing this character. This emphasizes that, even though the noise impact from a data quality perspective may be isolated to particular characters, its impact on the classification output may be broader. Therefore, it is not sufficient to only monitor examples that were explicitly impacted by noise, but instead to have a holistic view of where performance drops to best understand ramifications. 

Next, we try to understand the effects on fraud detection. For simplicity, if we assume that the error rate of each character is constant for all fonts and requires at least one misrecognition for the blacklist expression to fail, we can calculate an error likelihood for specific word inputs based on character-based accuracy:
\begin{equation}
\begin{split}
    \text{ Error(word)} = \text{Error($c_1, c_2, ... ,c_n$)}  
    = 1 - \prod_{i=1}^{n} \text{Accuracy}(c_i)
    \end{split}
\end{equation}
where $\text{Accuracy}(c)$ is the model accuracy on character $c$. Thus, if the financial services team formed a blacklist of particular words to catch, then we can estimate the increase in error for each blacklisted word when the OCR model updates to $h_2$ (Table \ref{tab:pipeline_downstream_errors}) by knowing the model accuracy for each character. Additionally, because of the nature of label noise, other characters beyond the manipulated targets shift in performance, such as the letter "Z", causing an increase in error for all words in the blacklist that account for misrecognitions of "Z" (e.g., "Zup" instead of "7up" drink). Thus, while the overall accuracy of word recognition might improve after the update, the performance of the system on the specific words that are included in the blacklist heuristics may degrade significantly. 

\begin{table}[t]
  \caption{Incompatible OCR examples (25\% noise).}
  \label{tab:ocr_examples}
  \footnotesize
  \centering
  \tabcolsep = 0.005\textwidth
  \begin{tabular}{ccc}
    \toprule
    \includegraphics[width=0.4cm]{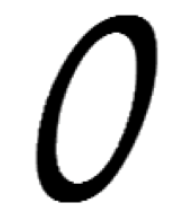}  & \includegraphics[width=0.49cm]{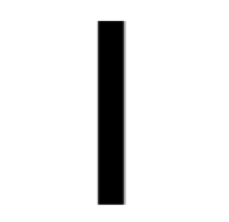} & \includegraphics[width=0.45cm]{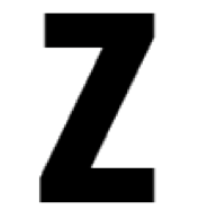}\\
    \midrule
    True Label: Digit 0 & True Label: lower L & True Label: upper Z \\
    19\% of incompatible points & 16\% of incompatible points & 13\% of incompatible points\\ \hline
    Model 1 Accuracy: 89\% & Model 1 Accuracy: 77\% & Model 1 Accuracy: 79\%\\
    Model 2 Accuracy: 10\% & Model 2 Accuracy: 17\% & Model 2 Accuracy: 21\%\\
  \bottomrule
\end{tabular}
\end{table}

\begin{table}[t]
  \caption{Downstream failures in receipt fraud detection.}
  \footnotesize
  \label{tab:pipeline_downstream_errors}
  \centering
  \tabcolsep = 0.005\textwidth
  \label{tab:attacks}
  \begin{tabular}{lll}
    \toprule
    Fraud Attack & Error Score ($h_1$)  & Error Score ($h_2$)\\
    \midrule
    "Nlke" & 55\% & 86.4\% \\
    "G00gle" & 85.3\% & 99.01\%\\
    "ZUP" & 49\% & 84.6\%\\
  \bottomrule
\end{tabular}
\end{table}

Without considering backward compatibility as captured by the \textbf{BTC} and \textbf{BEC} scores, the accuracy gain of the OCR system may be enough of an incentive for its designers to update the model and unintentionally cause significant losses for the fraud detection team. Note that, since the error score definition is conservative (i.e., only one character needs to be wrong for fraud detection to fail) and the average accuracy rates of $h_1$ and $h_2$ classifiers (currently 64\% and 77\%) could be improved with more advanced architectures, the issue would still occur even for more accurate classifiers.

Notably, there may be updates of an OCR system that do not involve noise, yet harm backward compatibility. For example, the OCR system could leverage natural language estimates on word likelihoods before a prediction---thus, assuming "G00gle" would more likely be "Google". The assumption that this will improve the model will cause serious issues for downstream tasks that rely on exactly detecting the unlikely words.

\section{Discussion \& Conclusion}
We showed that measuring backward compatibility can identify unreliability issues during an update and help ML practitioners avoid unexpected downstream failures. As ML pipelines designed for interaction with human users become larger and more pervasive, we expect backward compatibility to become an important property for assuring performance and reusability of models over time, and for maintaining trust with end users. While significant efforts continue with building end-to-end differentiable systems, we expect composable, modular systems to remain relevant per interpretability needs, ease of error detection, and worst-case performance analysis in safety-critical systems~\cite{karpathySoftware20}.
Examples of such complex systems include: 
(1) search engines relying on sensitive metrics for different demographics and markets during page ranking; (2) motion sensing in video game hardware that rely on hybrid data and physics-based models for detection and tracking; and, (3) productivity applications that combine deterministic software with statistical models for delivering consistent interaction with users.

We note that lack of backward compatibility may not  translate to a drop in performance in the real-world. Benchmark datasets used for evaluation may contain examples that are rarely encountered during future uses of the system, and if not related to a high-stakes situation, it might be desirable to sacrifice backward compatibility for accuracy gains. Additionally, user expectations or task definitions in the evolving world may change, which might make evaluation benchmarks obsolete. 

As a future direction, the monitoring of backward compatibility would benefit from stronger explanation and visualization techniques. Although error analysis approaches~\cite{nushi2018towards,zhang2018manifold} provide reports for single-model performance by identifying explainable regions of data more likely to fail, such visualizations comparing multiple models should be developed. Furthermore, current approaches to addressing issues such as data noise are often model agnostic, and assume that the model is always learned from scratch. Real-world deployment requirements can be better met if data denoising and repair is informed by the previous model that was trained on an initial, cleaner dataset.  Finally, we investigate simple noise settings to reflect practices with benign intentions where the practitioner decides to integrate larger and richer data that might contain noise. An interesting future direction would be to study the effect of adversarial training attacks on backward compatibility.

We hope that our work highlights the value of moving beyond evaluations that center on the aggregate performance of models in isolation and traditional assumptions of uniform monotonic improvements. Engineering of ML systems will often require careful approaches to model and data versioning that include methods for identifying and addressing challenges of backward compatibility.

\bibliographystyle{ACM-Reference-Format}
\bibliography{sample-base}


\begin{thebibliography}{42}

%%% the Web address.

\ifx \showCODEN    \undefined \def \showCODEN     #1{\unskip}     \fi
\ifx \showDOI      \undefined \def \showDOI       #1{#1}\fi
\ifx \showISBNx    \undefined \def \showISBNx     #1{\unskip}     \fi
\ifx \showISBNxiii \undefined \def \showISBNxiii  #1{\unskip}     \fi
\ifx \showISSN     \undefined \def \showISSN      #1{\unskip}     \fi
\ifx \showLCCN     \undefined \def \showLCCN      #1{\unskip}     \fi
\ifx \shownote     \undefined \def \shownote      #1{#1}          \fi
\ifx \showarticletitle \undefined \def \showarticletitle #1{#1}   \fi
\ifx \showURL      \undefined \def \showURL       {\relax}        \fi
% invisible to TeX
\providecommand\bibfield[2]{#2}
\providecommand\bibinfo[2]{#2}
\providecommand\natexlab[1]{#1}
\providecommand\showeprint[2][]{arXiv:#2}

\bibitem[\protect\citeauthoryear{Amershi, Begel, Bird, DeLine, Gall, Kamar,
  Nagappan, Nushi, and Zimmermann}{Amershi et~al\mbox{.}}{2019}]%
        {amershi2019software}
\bibfield{author}{\bibinfo{person}{Saleema Amershi}, \bibinfo{person}{Andrew
  Begel}, \bibinfo{person}{Christian Bird}, \bibinfo{person}{Robert DeLine},
  \bibinfo{person}{Harald Gall}, \bibinfo{person}{Ece Kamar},
  \bibinfo{person}{Nachiappan Nagappan}, \bibinfo{person}{Besmira Nushi}, {and}
  \bibinfo{person}{Thomas Zimmermann}.} \bibinfo{year}{2019}\natexlab{}.
\newblock \showarticletitle{Software engineering for machine learning: A case
  study}. In \bibinfo{booktitle}{\emph{ICSE-SEIP}}. IEEE.
\newblock


\bibitem[\protect\citeauthoryear{Andrist, Bohus, Kamar, and Horvitz}{Andrist
  et~al\mbox{.}}{2017}]%
        {andrist2017went}
\bibfield{author}{\bibinfo{person}{Sean Andrist}, \bibinfo{person}{Dan Bohus},
  \bibinfo{person}{Ece Kamar}, {and} \bibinfo{person}{Eric Horvitz}.}
  \bibinfo{year}{2017}\natexlab{}.
\newblock \showarticletitle{What went wrong and why? diagnosing situated
  interaction failures in the wild}. In \bibinfo{booktitle}{\emph{ICSR}}.
\newblock


\bibitem[\protect\citeauthoryear{Bansal, Nushi, Kamar, Weld, Lasecki, and
  Horvitz}{Bansal et~al\mbox{.}}{2019}]%
        {bansal2019updates}
\bibfield{author}{\bibinfo{person}{Gagan Bansal}, \bibinfo{person}{Besmira
  Nushi}, \bibinfo{person}{Ece Kamar}, \bibinfo{person}{Daniel~S Weld},
  \bibinfo{person}{Walter~S Lasecki}, {and} \bibinfo{person}{Eric Horvitz}.}
  \bibinfo{year}{2019}\natexlab{}.
\newblock \showarticletitle{Updates in human-ai teams: Understanding and
  addressing the performance/compatibility tradeoff}. In
  \bibinfo{booktitle}{\emph{AAAI}}, Vol.~\bibinfo{volume}{33}.
  \bibinfo{pages}{2429--2437}.
\newblock


\bibitem[\protect\citeauthoryear{Bosch}{Bosch}{2009}]%
        {bosch2009software}
\bibfield{author}{\bibinfo{person}{Jan Bosch}.}
  \bibinfo{year}{2009}\natexlab{}.
\newblock \showarticletitle{From software product lines to software
  ecosystems}. In \bibinfo{booktitle}{\emph{SPLC}}.
\newblock


\bibitem[\protect\citeauthoryear{Cheplygina, de~Bruijne, and Pluim}{Cheplygina
  et~al\mbox{.}}{2019}]%
        {cheplygina2019not}
\bibfield{author}{\bibinfo{person}{Veronika Cheplygina},
  \bibinfo{person}{Marleen de Bruijne}, {and} \bibinfo{person}{Josien~PW
  Pluim}.} \bibinfo{year}{2019}\natexlab{}.
\newblock \showarticletitle{Not-so-supervised: a survey of semi-supervised,
  multi-instance, and transfer learning in medical image analysis}.
\newblock \bibinfo{journal}{\emph{Medical image analysis}}
  \bibinfo{volume}{54} (\bibinfo{year}{2019}), \bibinfo{pages}{280--296}.
\newblock


\bibitem[\protect\citeauthoryear{Cheplygina, Peña, Pederson, Lynch, Sørensen,
  and de~Bruijne}{Cheplygina et~al\mbox{.}}{2018}]%
        {Cheplygina2018Copd}
\bibfield{author}{\bibinfo{person}{Veronika Cheplygina},
  \bibinfo{person}{Isabel~Pino Peña}, \bibinfo{person}{Jesper~Holst Pederson},
  \bibinfo{person}{David Lynch}, \bibinfo{person}{Lauge Sørensen}, {and}
  \bibinfo{person}{Marleen de Bruijne}.} \bibinfo{year}{2018}\natexlab{}.
\newblock \showarticletitle{Transfer Learning for Multicenter Classification of
  Chronic Obstructive Pulmonary Disease}.
\newblock \bibinfo{journal}{\emph{IEEE Journal of Biomedical and Health
  Informatics}} \bibinfo{volume}{22}, \bibinfo{number}{5}
  (\bibinfo{year}{2018}), \bibinfo{pages}{1486--1496}.
\newblock


\bibitem[\protect\citeauthoryear{Chung, Kraska, Polyzotis, Tae, and
  Whang}{Chung et~al\mbox{.}}{2019}]%
        {chung2019slice}
\bibfield{author}{\bibinfo{person}{Yeounoh Chung}, \bibinfo{person}{Tim
  Kraska}, \bibinfo{person}{Neoklis Polyzotis}, \bibinfo{person}{Ki~Hyun Tae},
  {and} \bibinfo{person}{Steven~Euijong Whang}.}
  \bibinfo{year}{2019}\natexlab{}.
\newblock \showarticletitle{Slice finder: Automated data slicing for model
  validation}. In \bibinfo{booktitle}{\emph{ICDE}}.
\newblock


\bibitem[\protect\citeauthoryear{Dallachiesa, Ebaid, Eldawy, Elmagarmid, Ilyas,
  Ouzzani, and Tang}{Dallachiesa et~al\mbox{.}}{2013}]%
        {DBLP:conf/sigmod/DallachiesaEEEIOT13}
\bibfield{author}{\bibinfo{person}{Michele Dallachiesa}, \bibinfo{person}{Amr
  Ebaid}, \bibinfo{person}{Ahmed Eldawy}, \bibinfo{person}{Ahmed~K.
  Elmagarmid}, \bibinfo{person}{Ihab~F. Ilyas}, \bibinfo{person}{Mourad
  Ouzzani}, {and} \bibinfo{person}{Nan Tang}.} \bibinfo{year}{2013}\natexlab{}.
\newblock \showarticletitle{{NADEEF:} a commodity data cleaning system}. In
  \bibinfo{booktitle}{\emph{SIGMOD}},
  \bibfield{editor}{\bibinfo{person}{Kenneth~A. Ross}, \bibinfo{person}{Divesh
  Srivastava}, {and} \bibinfo{person}{Dimitris Papadias}} (Eds.).
  \bibinfo{publisher}{{ACM}}, \bibinfo{pages}{541--552}.
\newblock
\urldef\tempurl%
\url{https://doi.org/10.1145/2463676.2465327}
\showDOI{\tempurl}


\bibitem[\protect\citeauthoryear{de~Campos, Babu, and Varma}{de~Campos
  et~al\mbox{.}}{2009}]%
        {deCampos09}
\bibfield{author}{\bibinfo{person}{T.~E. de Campos}, \bibinfo{person}{B.~R.
  Babu}, {and} \bibinfo{person}{M. Varma}.} \bibinfo{year}{2009}\natexlab{}.
\newblock \showarticletitle{Character recognition in natural images}. In
  \bibinfo{booktitle}{\emph{VISAPP}}.
\newblock


\bibitem[\protect\citeauthoryear{FICO}{FICO}{2020}]%
        {fico2020data}
\bibfield{author}{\bibinfo{person}{FICO}.} \bibinfo{year}{2018 (accessed
  February 13, 2020)}\natexlab{}.
\newblock \bibinfo{booktitle}{\emph{Explainable machine learning challenge.}}
\newblock
\newblock
\shownote{\url{https://community.fico.com/s/explainable-machine-learning-challenge?tabset-3158a=4fbc8}.}


\bibitem[\protect\citeauthoryear{Fr{\'e}nay and Verleysen}{Fr{\'e}nay and
  Verleysen}{2013}]%
        {frenay2013classification}
\bibfield{author}{\bibinfo{person}{Beno{\^\i}t Fr{\'e}nay} {and}
  \bibinfo{person}{Michel Verleysen}.} \bibinfo{year}{2013}\natexlab{}.
\newblock \showarticletitle{Classification in the presence of label noise: a
  survey}.
\newblock \bibinfo{journal}{\emph{IEEE transactions on neural networks and
  learning systems}} \bibinfo{volume}{25}, \bibinfo{number}{5}
  (\bibinfo{year}{2013}), \bibinfo{pages}{845--869}.
\newblock


\bibitem[\protect\citeauthoryear{French}{French}{1999}]%
        {french1999catastrophic}
\bibfield{author}{\bibinfo{person}{Robert~M French}.}
  \bibinfo{year}{1999}\natexlab{}.
\newblock \showarticletitle{Catastrophic forgetting in connectionist networks}.
\newblock \bibinfo{journal}{\emph{Trends in cognitive sciences}}
  \bibinfo{volume}{3}, \bibinfo{number}{4} (\bibinfo{year}{1999}),
  \bibinfo{pages}{128--135}.
\newblock


\bibitem[\protect\citeauthoryear{Goodfellow, Mirza, Xiao, Courville, and
  Bengio}{Goodfellow et~al\mbox{.}}{2013}]%
        {goodfellow2013empirical}
\bibfield{author}{\bibinfo{person}{Ian~J Goodfellow}, \bibinfo{person}{Mehdi
  Mirza}, \bibinfo{person}{Da Xiao}, \bibinfo{person}{Aaron Courville}, {and}
  \bibinfo{person}{Yoshua Bengio}.} \bibinfo{year}{2013}\natexlab{}.
\newblock \showarticletitle{An empirical investigation of catastrophic
  forgetting in gradient-based neural networks}.
\newblock \bibinfo{journal}{\emph{arXiv preprint arXiv:1312.6211}}
  (\bibinfo{year}{2013}).
\newblock


\bibitem[\protect\citeauthoryear{Hendrycks and Dietterich}{Hendrycks and
  Dietterich}{2019}]%
        {hendrycks2019benchmarking}
\bibfield{author}{\bibinfo{person}{Dan Hendrycks} {and} \bibinfo{person}{Thomas
  Dietterich}.} \bibinfo{year}{2019}\natexlab{}.
\newblock \showarticletitle{Benchmarking neural network robustness to common
  corruptions and perturbations}.
\newblock \bibinfo{journal}{\emph{ICLR}}.
\newblock


\bibitem[\protect\citeauthoryear{Holmstrom and Koistinen}{Holmstrom and
  Koistinen}{1992}]%
        {holmstrom1992noise}
\bibfield{author}{\bibinfo{person}{Lasse Holmstrom} {and}
  \bibinfo{person}{Petri Koistinen}.} \bibinfo{year}{1992}\natexlab{}.
\newblock \showarticletitle{Using additive noise in back-propagation training}.
\newblock \bibinfo{journal}{\emph{IEEE Transactions on Neural Networks}}
  \bibinfo{volume}{3} (\bibinfo{year}{1992}).
\newblock


\bibitem[\protect\citeauthoryear{Jiang, Zhou, Leung, Li, and Fei-Fei}{Jiang
  et~al\mbox{.}}{2017}]%
        {Jiang2017MentorNetLD}
\bibfield{author}{\bibinfo{person}{Lu Jiang}, \bibinfo{person}{Zhengyuan Zhou},
  \bibinfo{person}{Thomas Leung}, \bibinfo{person}{Li-Jia Li}, {and}
  \bibinfo{person}{Li Fei-Fei}.} \bibinfo{year}{2017}\natexlab{}.
\newblock \showarticletitle{MentorNet: Learning Data-Driven Curriculum for Very
  Deep Neural Networks on Corrupted Labels}. In
  \bibinfo{booktitle}{\emph{ICML}}.
\newblock


\bibitem[\protect\citeauthoryear{Karpathy}{Karpathy}{2017}]%
        {karpathySoftware20}
\bibfield{author}{\bibinfo{person}{Andrej Karpathy}.}
  \bibinfo{year}{2017}\natexlab{}.
\newblock \bibinfo{title}{Software 2.0}.
\newblock
\newblock
\urldef\tempurl%
\url{https://medium.com/@karpathy/software-2-0-a64152b37c35}
\showURL{%
\tempurl}


\bibitem[\protect\citeauthoryear{Kemker, McClure, Abitino, Hayes, and
  Kanan}{Kemker et~al\mbox{.}}{2018}]%
        {kemker2018measuring}
\bibfield{author}{\bibinfo{person}{Ronald Kemker}, \bibinfo{person}{Marc
  McClure}, \bibinfo{person}{Angelina Abitino}, \bibinfo{person}{Tyler~L
  Hayes}, {and} \bibinfo{person}{Christopher Kanan}.}
  \bibinfo{year}{2018}\natexlab{}.
\newblock \showarticletitle{Measuring catastrophic forgetting in neural
  networks}. In \bibinfo{booktitle}{\emph{AAAI}}.
\newblock


\bibitem[\protect\citeauthoryear{Krause, Sapp, Howard, Zhou, Toshev, Duerig,
  Philbin, and Fei-Fei}{Krause et~al\mbox{.}}{2016}]%
        {krause2016unreasonable}
\bibfield{author}{\bibinfo{person}{Jonathan Krause}, \bibinfo{person}{Benjamin
  Sapp}, \bibinfo{person}{Andrew Howard}, \bibinfo{person}{Howard Zhou},
  \bibinfo{person}{Alexander Toshev}, \bibinfo{person}{Tom Duerig},
  \bibinfo{person}{James Philbin}, {and} \bibinfo{person}{Li Fei-Fei}.}
  \bibinfo{year}{2016}\natexlab{}.
\newblock \showarticletitle{The unreasonable effectiveness of noisy data for
  fine-grained recognition}. In \bibinfo{booktitle}{\emph{ECCV}}.
\newblock


\bibitem[\protect\citeauthoryear{Krizhevsky}{Krizhevsky}{2009}]%
        {Krizhevsky09learningmultiple}
\bibfield{author}{\bibinfo{person}{Alex Krizhevsky}.}
  \bibinfo{year}{2009}\natexlab{}.
\newblock \showarticletitle{Learning multiple layers of features from tiny
  images}.
\newblock  (\bibinfo{year}{2009}).
\newblock


\bibitem[\protect\citeauthoryear{Kruppa and Prakash}{Kruppa and
  Prakash}{2008}]%
        {kruppa2008patent}
\bibfield{author}{\bibinfo{person}{Robert~William Kruppa} {and}
  \bibinfo{person}{Ravinder Prakash}.} \bibinfo{year}{U.S. Patent 0 298 668,
  Dec. 2008}\natexlab{}.
\newblock \bibinfo{title}{Method for fraud detection using multiple scan
  technologies}.
\newblock
\newblock


\bibitem[\protect\citeauthoryear{Kurakin, Goodfellow, and Bengio}{Kurakin
  et~al\mbox{.}}{2017}]%
        {Kurakin2016AdversarialML}
\bibfield{author}{\bibinfo{person}{Alexey Kurakin}, \bibinfo{person}{Ian~J.
  Goodfellow}, {and} \bibinfo{person}{Samy Bengio}.}
  \bibinfo{year}{2017}\natexlab{}.
\newblock \showarticletitle{Adversarial Machine Learning at Scale}.
\newblock \bibinfo{journal}{\emph{ICLR}}.
\newblock


\bibitem[\protect\citeauthoryear{Law and Ahn}{Law and Ahn}{2011}]%
        {law2011human}
\bibfield{author}{\bibinfo{person}{Edith Law} {and} \bibinfo{person}{Luis~von
  Ahn}.} \bibinfo{year}{2011}\natexlab{}.
\newblock \showarticletitle{Human computation}.
\newblock \bibinfo{journal}{\emph{Synthesis lectures on artificial intelligence
  and machine learning}} \bibinfo{volume}{5}, \bibinfo{number}{3}
  (\bibinfo{year}{2011}), \bibinfo{pages}{1--121}.
\newblock


\bibitem[\protect\citeauthoryear{LeCun, Cortes, and Burges}{LeCun
  et~al\mbox{.}}{2010}]%
        {lecun2010mnist}
\bibfield{author}{\bibinfo{person}{Yann LeCun}, \bibinfo{person}{Corinna
  Cortes}, {and} \bibinfo{person}{CJ Burges}.} \bibinfo{year}{2010}\natexlab{}.
\newblock \showarticletitle{MNIST handwritten digit database}.
\newblock \bibinfo{journal}{\emph{ATT Labs [Online]. Available:
  http://yann.lecun.com/exdb/mnist}}  \bibinfo{volume}{2}
  (\bibinfo{year}{2010}).
\newblock


\bibitem[\protect\citeauthoryear{Li, Yang, Song, Cao, Luo, and Li}{Li
  et~al\mbox{.}}{2017}]%
        {Li2017LearningFN}
\bibfield{author}{\bibinfo{person}{Yuncheng Li}, \bibinfo{person}{Jianchao
  Yang}, \bibinfo{person}{Yale Song}, \bibinfo{person}{Liangliang Cao},
  \bibinfo{person}{Jiebo Luo}, {and} \bibinfo{person}{Li-Jia Li}.}
  \bibinfo{year}{2017}\natexlab{}.
\newblock \showarticletitle{Learning from Noisy Labels with Distillation}.
\newblock \bibinfo{journal}{\emph{ICCV}}, \bibinfo{pages}{1928--1936}.
\newblock


\bibitem[\protect\citeauthoryear{Maas, Daly, Pham, Huang, Ng, and Potts}{Maas
  et~al\mbox{.}}{2011}]%
        {maas2011imdb}
\bibfield{author}{\bibinfo{person}{Andrew~L. Maas}, \bibinfo{person}{Raymond~E.
  Daly}, \bibinfo{person}{Peter~T. Pham}, \bibinfo{person}{Dan Huang},
  \bibinfo{person}{Andrew~Y. Ng}, {and} \bibinfo{person}{Christopher Potts}.}
  \bibinfo{year}{2011}\natexlab{}.
\newblock \showarticletitle{Learning Word Vectors for Sentiment Analysis}. In
  \bibinfo{booktitle}{\emph{ACL}}.
\newblock


\bibitem[\protect\citeauthoryear{Madry, Makelov, Schmidt, Tsipras, and
  Vladu}{Madry et~al\mbox{.}}{2017}]%
        {Madry2017TowardsDL}
\bibfield{author}{\bibinfo{person}{Aleksander Madry},
  \bibinfo{person}{Aleksandar Makelov}, \bibinfo{person}{Ludwig Schmidt},
  \bibinfo{person}{Dimitris Tsipras}, {and} \bibinfo{person}{Adrian Vladu}.}
  \bibinfo{year}{2017}\natexlab{}.
\newblock \showarticletitle{Towards Deep Learning Models Resistant to
  Adversarial Attacks}.
\newblock \bibinfo{journal}{\emph{ICLR}}.
\newblock


\bibitem[\protect\citeauthoryear{McCloskey and Cohen}{McCloskey and
  Cohen}{1989}]%
        {mccloskey1989catastrophic}
\bibfield{author}{\bibinfo{person}{Michael McCloskey} {and}
  \bibinfo{person}{Neal~J Cohen}.} \bibinfo{year}{1989}\natexlab{}.
\newblock \showarticletitle{Catastrophic interference in connectionist
  networks: The sequential learning problem}.
\newblock In \bibinfo{booktitle}{\emph{Psychology of learning and motivation}}.
  Vol.~\bibinfo{volume}{24}. \bibinfo{publisher}{Elsevier},
  \bibinfo{pages}{109--165}.
\newblock


\bibitem[\protect\citeauthoryear{Mendels, Cooper, Soto, Hirschberg, Gales,
  Knill, Ragni, and Wang}{Mendels et~al\mbox{.}}{2015}]%
        {Mendels2015Speech}
\bibfield{author}{\bibinfo{person}{Gideon Mendels}, \bibinfo{person}{Erica
  Cooper}, \bibinfo{person}{Victor Soto}, \bibinfo{person}{Julia Hirschberg},
  \bibinfo{person}{Mark Gales}, \bibinfo{person}{Kate Knill},
  \bibinfo{person}{Anton Ragni}, {and} \bibinfo{person}{Haipeng Wang}.}
  \bibinfo{year}{2015}\natexlab{}.
\newblock \showarticletitle{Improving Speech Recognition and Keyword Search for
  Low Resource Languages Using Web Data}.
\newblock \bibinfo{journal}{\emph{ISCA}} (\bibinfo{year}{2015}).
\newblock


\bibitem[\protect\citeauthoryear{Natarajan, Dhillon, Ravikumar, and
  Tewari}{Natarajan et~al\mbox{.}}{2013}]%
        {Natarajan2013LearningWN}
\bibfield{author}{\bibinfo{person}{Nagarajan Natarajan},
  \bibinfo{person}{Inderjit~S. Dhillon}, \bibinfo{person}{Pradeep Ravikumar},
  {and} \bibinfo{person}{Ambuj Tewari}.} \bibinfo{year}{2013}\natexlab{}.
\newblock \showarticletitle{Learning with Noisy Labels}. In
  \bibinfo{booktitle}{\emph{NeurIPS}}.
\newblock


\bibitem[\protect\citeauthoryear{Nushi, Kamar, and Horvitz}{Nushi
  et~al\mbox{.}}{2018}]%
        {nushi2018towards}
\bibfield{author}{\bibinfo{person}{Besmira Nushi}, \bibinfo{person}{Ece Kamar},
  {and} \bibinfo{person}{Eric Horvitz}.} \bibinfo{year}{2018}\natexlab{}.
\newblock \showarticletitle{Towards accountable ai: Hybrid human-machine
  analyses for characterizing system failure}. In
  \bibinfo{booktitle}{\emph{HCOMP}}.
\newblock


\bibitem[\protect\citeauthoryear{Nushi, Kamar, Horvitz, and Kossmann}{Nushi
  et~al\mbox{.}}{2017}]%
        {nushi2017human}
\bibfield{author}{\bibinfo{person}{Besmira Nushi}, \bibinfo{person}{Ece Kamar},
  \bibinfo{person}{Eric Horvitz}, {and} \bibinfo{person}{Donald Kossmann}.}
  \bibinfo{year}{2017}\natexlab{}.
\newblock \showarticletitle{On human intellect and machine failures:
  Troubleshooting integrative machine learning systems}. In
  \bibinfo{booktitle}{\emph{AAAI}}.
\newblock


\bibitem[\protect\citeauthoryear{Rahm and Do}{Rahm and Do}{2000}]%
        {rahm2000data}
\bibfield{author}{\bibinfo{person}{Erhard Rahm} {and} \bibinfo{person}{Hong~Hai
  Do}.} \bibinfo{year}{2000}\natexlab{}.
\newblock \showarticletitle{Data cleaning: Problems and current approaches}.
\newblock \bibinfo{journal}{\emph{IEEE Data Eng. Bull.}} \bibinfo{volume}{23},
  \bibinfo{number}{4} (\bibinfo{year}{2000}), \bibinfo{pages}{3--13}.
\newblock


\bibitem[\protect\citeauthoryear{Ratner, Bach, Ehrenberg, Fries, Wu, and
  R{\'e}}{Ratner et~al\mbox{.}}{2019}]%
        {ratner2019snorkel}
\bibfield{author}{\bibinfo{person}{Alexander Ratner},
  \bibinfo{person}{Stephen~H Bach}, \bibinfo{person}{Henry Ehrenberg},
  \bibinfo{person}{Jason Fries}, \bibinfo{person}{Sen Wu}, {and}
  \bibinfo{person}{Christopher R{\'e}}.} \bibinfo{year}{2019}\natexlab{}.
\newblock \showarticletitle{Snorkel: Rapid training data creation with weak
  supervision}.
\newblock \bibinfo{journal}{\emph{The VLDB Journal}} (\bibinfo{year}{2019}),
  \bibinfo{pages}{1--22}.
\newblock


\bibitem[\protect\citeauthoryear{Rekatsinas, Chu, Ilyas, and
  R{\'{e}}}{Rekatsinas et~al\mbox{.}}{2017}]%
        {DBLP:journals/pvldb/RekatsinasCIR17}
\bibfield{author}{\bibinfo{person}{Theodoros Rekatsinas}, \bibinfo{person}{Xu
  Chu}, \bibinfo{person}{Ihab~F. Ilyas}, {and} \bibinfo{person}{Christopher
  R{\'{e}}}.} \bibinfo{year}{2017}\natexlab{}.
\newblock \showarticletitle{HoloClean: Holistic Data Repairs with Probabilistic
  Inference}.
\newblock \bibinfo{journal}{\emph{{PVLDB}}} \bibinfo{volume}{10},
  \bibinfo{number}{11} (\bibinfo{year}{2017}), \bibinfo{pages}{1190--1201}.
\newblock
\urldef\tempurl%
\url{https://doi.org/10.14778/3137628.3137631}
\showDOI{\tempurl}


\bibitem[\protect\citeauthoryear{Sculley, Holt, Golovin, Davydov, Phillips,
  Ebner, Chaudhary, Young, Crespo, and Dennison}{Sculley et~al\mbox{.}}{2015}]%
        {sculley2015hidden}
\bibfield{author}{\bibinfo{person}{David Sculley}, \bibinfo{person}{Gary Holt},
  \bibinfo{person}{Daniel Golovin}, \bibinfo{person}{Eugene Davydov},
  \bibinfo{person}{Todd Phillips}, \bibinfo{person}{Dietmar Ebner},
  \bibinfo{person}{Vinay Chaudhary}, \bibinfo{person}{Michael Young},
  \bibinfo{person}{Jean-Francois Crespo}, {and} \bibinfo{person}{Dan
  Dennison}.} \bibinfo{year}{2015}\natexlab{}.
\newblock \showarticletitle{Hidden technical debt in machine learning systems}.
  In \bibinfo{booktitle}{\emph{NeurIPS}}.
\newblock


\bibitem[\protect\citeauthoryear{Sukhbaatar, Bruna, Paluri, Bourdev, and
  Fergus}{Sukhbaatar et~al\mbox{.}}{2015}]%
        {Sukhbaatar2014TrainingCN}
\bibfield{author}{\bibinfo{person}{Sainbayar Sukhbaatar}, \bibinfo{person}{Joan
  Bruna}, \bibinfo{person}{Manohar Paluri}, \bibinfo{person}{Lubomir~D.
  Bourdev}, {and} \bibinfo{person}{Rob Fergus}.}
  \bibinfo{year}{2015}\natexlab{}.
\newblock \showarticletitle{Training Convolutional Networks with Noisy Labels}.
  In \bibinfo{booktitle}{\emph{ICLR 2015}}.
\newblock


\bibitem[\protect\citeauthoryear{Toneva, Sordoni, Combes, Trischler, Bengio,
  and Gordon}{Toneva et~al\mbox{.}}{2019}]%
        {toneva2018empirical}
\bibfield{author}{\bibinfo{person}{Mariya Toneva}, \bibinfo{person}{Alessandro
  Sordoni}, \bibinfo{person}{Remi Tachet~des Combes}, \bibinfo{person}{Adam
  Trischler}, \bibinfo{person}{Yoshua Bengio}, {and}
  \bibinfo{person}{Geoffrey~J Gordon}.} \bibinfo{year}{2019}\natexlab{}.
\newblock \showarticletitle{An empirical study of example forgetting during
  deep neural network learning}.
\newblock \bibinfo{journal}{\emph{ICLR}}.
\newblock


\bibitem[\protect\citeauthoryear{Von~Ahn, Blum, Hopper, and Langford}{Von~Ahn
  et~al\mbox{.}}{2003}]%
        {von2003captcha}
\bibfield{author}{\bibinfo{person}{Luis Von~Ahn}, \bibinfo{person}{Manuel
  Blum}, \bibinfo{person}{Nicholas~J Hopper}, {and} \bibinfo{person}{John
  Langford}.} \bibinfo{year}{2003}\natexlab{}.
\newblock \showarticletitle{CAPTCHA: Using hard AI problems for security}. In
  \bibinfo{booktitle}{\emph{EUROCRYPT}}. Springer.
\newblock


\bibitem[\protect\citeauthoryear{Wen, Yu, and Greiner}{Wen
  et~al\mbox{.}}{2014}]%
        {Wen2014Test}
\bibfield{author}{\bibinfo{person}{Junfeng Wen}, \bibinfo{person}{Chun-Nam Yu},
  {and} \bibinfo{person}{Russell Greiner}.} \bibinfo{year}{2014}\natexlab{}.
\newblock \showarticletitle{Robust Learning under Uncertain Test Distributions:
  Relating Covariate Shift to Model Misspecification}.
\newblock \bibinfo{journal}{\emph{ICML}}.
\newblock


\bibitem[\protect\citeauthoryear{Zhang, Wang, Molino, Li, and Ebert}{Zhang
  et~al\mbox{.}}{2018}]%
        {zhang2018manifold}
\bibfield{author}{\bibinfo{person}{Jiawei Zhang}, \bibinfo{person}{Yang Wang},
  \bibinfo{person}{Piero Molino}, \bibinfo{person}{Lezhi Li}, {and}
  \bibinfo{person}{David~S Ebert}.} \bibinfo{year}{2018}\natexlab{}.
\newblock \showarticletitle{Manifold: A model-agnostic framework for
  interpretation and diagnosis of machine learning models}.
\newblock \bibinfo{journal}{\emph{TVCG}} \bibinfo{volume}{25},
  \bibinfo{number}{1} (\bibinfo{year}{2018}), \bibinfo{pages}{364--373}.
\newblock


\bibitem[\protect\citeauthoryear{Zhang, Gao, Ma, Lyu, and Kim}{Zhang
  et~al\mbox{.}}{2019}]%
        {zhang2019empirical}
\bibfield{author}{\bibinfo{person}{Tianyi Zhang}, \bibinfo{person}{Cuiyun Gao},
  \bibinfo{person}{Lei Ma}, \bibinfo{person}{Michael~R Lyu}, {and}
  \bibinfo{person}{Miryung Kim}.} \bibinfo{year}{2019}\natexlab{}.
\newblock \showarticletitle{An empirical study of common challenges in
  developing deep learning applications}. In \bibinfo{booktitle}{\emph{ISSRE}}.
\newblock


\end{thebibliography}

\end{document}